\documentclass[table]{article}

\PassOptionsToPackage{numbers, compress}{natbib}
\usepackage[preprint]{neurips_2025}
\usepackage{array}
\usepackage{amssymb} 
\usepackage{float}
\usepackage{graphicx}
\usepackage{amsmath}
\usepackage[utf8]{inputenc} 
\usepackage[T1]{fontenc}    
\usepackage{hyperref}       
\usepackage{url}            
\usepackage{booktabs}       
\usepackage{amsfonts}       
\usepackage{nicefrac}       
\usepackage{microtype}      
\usepackage{xcolor}         
\usepackage{xpatch}
\usepackage{makecell}
\usepackage{graphicx}
\usepackage{xpatch}
\usepackage{algorithm}
\usepackage{algorithmic}
\usepackage{amssymb}
\usepackage{pifont}
\usepackage{tcolorbox}
\usepackage{xcolor}
\usepackage{multirow}

\newcommand{\xmark}{\ding{55}} 

\makeatletter
\xapptocmd{\NAT@bibsetnum}{\setlength{\leftmargin}{0pt}\setlength{\itemindent}{\labelwidth}\addtolength{\itemindent}{\labelsep}}{}{}
\makeatother

\title{SceneRAG: Scene-level Retrieval-Augmented Generation for Video Understanding}

\author{
\textbf{Nianbo Zeng}$^{1,2}$, 
\textbf{Haowen Hou}$^{1}$, 
\textbf{Fei Richard Yu}$^{1,2}$, 
\textbf{Si Shi}$^{1}$, 
\textbf{Ying Tiffany He}$^{2}$ \\
$^{1}$Guangdong Laboratory of Artificial Intelligence and Digital Economy (SZ), Shenzhen, China \\
$^{2}$College of Computer Science and Software Engineering, Shenzhen University, China \\
\texttt{\{zengnianbo, houhaowen, yufei, shisi\}@gml.ac.cn}, 
\texttt{tiffanyhe@szu.edu.cn}
}

\begin{document}

\maketitle

\begin{abstract}
Despite recent advances in retrieval-augmented generation (RAG) for video understanding, effectively understanding long-form video content remains underexplored due to the vast scale and high complexity of video data. Current RAG approaches typically segment videos into fixed-length chunks, which often disrupts the continuity of contextual information and fails to capture authentic scene boundaries. Inspired by the human ability to naturally organize continuous experiences into coherent scenes, we present SceneRAG, a unified framework that leverages large language models to segment videos into narrative-consistent scenes by processing ASR transcripts alongside temporal metadata. SceneRAG further sharpens these initial boundaries through lightweight heuristics and iterative correction. For each scene, the framework fuses information from both visual and textual modalities to extract entity relations and dynamically builds a knowledge graph, enabling robust multi-hop retrieval and generation that account for long-range dependencies. Experiments on the LongerVideos benchmark, featuring over 134 hours of diverse content, confirm that SceneRAG substantially outperforms prior baselines, achieving a win rate of up to 72.5 percent on generation tasks.

\end{abstract}

\section{Introduction}

Video has become the dominant medium for communication, education, and entertainment in the digital era. Unlike unimodal text or audio, video integrates visual frames, spoken language, on-screen graphics, and ambient sound into a continuous multimodal stream. This richness enables deep storytelling and immersive learning, but also produces massive, unstructured data that challenges both retrieval and generative models~\cite{abootorabi2025ask, yuan2025momentseeker, jin2023diffusionret}. In particular, static text-centric Retrieval-Augmented Generation (RAG) methods excel on documents but struggle to capture the temporal and multimodal complexity of video content~\cite{lewis2020retrieval,gao2023retrieval,jeong2025videorag}.

Existing video segmentation approaches~\cite{chen2024sharegpt4video,fan2024videoagent} often rely on fixed-length windows or naive sliding clips, which frequently misalign with true scene boundaries and yield fragmented narratives, ultimately degrading retrieval precision and downstream performance. Moreover, current RAG systems~\cite{asai2023self,lewis2020retrieval,chu2024graphrag} treat each segment as an isolated text block (e.g., subtitle snippet), ignoring inter-scene dependencies such as recurring characters, thematic motifs, and long-range references. Retrieval-augmented generation pipelines that process segments in isolation thus overlook critical narrative continuity, temporal coherence, and entity tracking.

A key insight to overcome these limitations comes from psychology~\cite{zacks2010brain}: humans naturally segment continuous experiences into discrete, meaningful “scenes” through an \emph{event segmentation} process. This ability enables us to efficiently organize, comprehend, and recall long sequences of events. Film editors, for example, leverage this principle through continuity editing, guiding viewers seamlessly across cuts and locations~\cite{cutting2014event, magliano2011impact, cutting2019large}. By breaking down long videos into self-contained, semantically coherent scenes—typically centered on a specific location, speaker, or topic—humans can more easily follow narratives and track salient entities and actions. Emulating this perceptual segmentation thus provides a principled foundation for automated video indexing, more precise retrieval, and robust cross-modal alignment in long-form video understanding.

\begin{figure}
\centering
\includegraphics[width=1\linewidth]{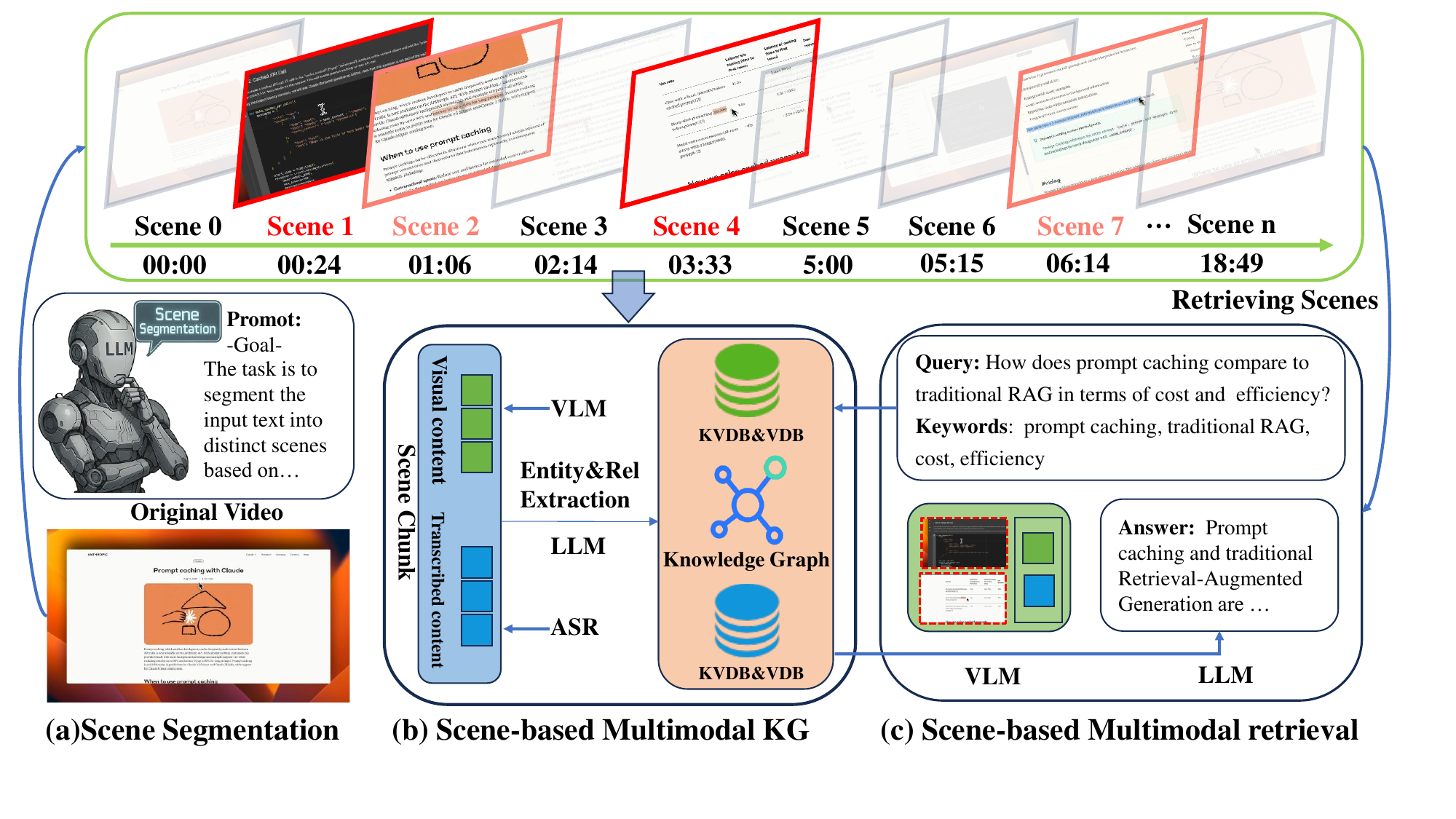}
\caption{\label{fig:frame}Overview of SceneRAG. Given a long video, SceneRAG segments it into scenes highlighted in green using an LLM and heuristics. For a query, relevant scenes are highlighted in red, and retrieved but irrelevant scenes are outlined in pink
through knowledge graph-based retrieva.}
\end{figure}

To address these limitations, we propose \textbf{SceneRAG}, a unified framework that begins with human-inspired, LLM-driven scene segmentation. By reasoning over transcripts, subtitles, and temporal cues, the LLM produces semantically coherent scene boundaries aligned with narrative flow~\cite{cheng2024videollama,zhang2023video}. To further enhance precision and robustness, we incorporate simple yet effective heuristics—such as Mute processing and segment alignment—along with a boundary correction mechanism to refine coarse boundaries and ensure accurate scene transition alignment. Next, SceneRAG constructs a lightweight knowledge graph~\cite{edge2024local} where nodes represent scenes, entities, and events, and edges capture semantic co-occurrence and temporal adjacency~\cite{kim2024scene}. By enabling multi-hop retrieval over this graph, the model recovers long-range dependencies that flat retrieval cannot model. Given a user query, SceneRAG retrieves a relevant subgraph, aggregates multimodal signals, and issues a unified prompt to an LLM, resulting in context-rich generation.

Our contributions are as follows.
\begin{itemize}
\item \textbf{Human-inspired Scene Segmentation.}We introduce an LLM-based algorithm that fuses vision, audio, and subtitle cues to generate semantically coherent scene boundaries, emulating cognitive event segmentation, thereby refining existing segmentation methods.
\item \textbf{Unified RAG Pipeline.} We propose a novel framework that cohesively combines dynamic segmentation, advanced graph construction, and retrieval-augmented generation, significantly enhancing the robustness and accuracy of long-form video understanding.

\item \textbf{Comprehensive Evaluation.} Experiments on public benchmarks—including Lecture, Documentary and Entertainment datasets—demonstrate SceneRAG’s substantial improvements in retrieval precision and generation quality compared to state-of-the-art baselines.
\end{itemize}

\section{Related work}

\paragraph{Multimodal Video Understanding as Foundation.}
Recent video-language models have advanced joint representations across vision, text, and audio, enabling more comprehensive video understanding. ViViT~\cite{arnab2021vivit} uses transformers for spatio-temporal classification, while Flamingo bridges pretrained vision and language models for multimodal inputs, including videos~\cite{alayrac2022flamingo}. MERLOT Reserve~\cite{zellers2022merlot} aligns video, subtitles, and audio to improve representation learning. VideoBERT~\cite{sun2019videobert} quantizes frames into “visual words” and uses a BERT-like model to capture high-level semantics and temporal dynamics for tasks like question answering. VideoAgent~\cite{wang2024videoagent} adds structured memory of events and objects, supporting complex LLM querying. These models provide strong multimodal encodings and inspire further exploration of temporal structure in video systems.

\paragraph{Retrieval-Augmented Generation.} 
With rapid advances in retrieval-augmented generation (RAG) frameworks, knowledge-enhanced generation methods have evolved considerably. RAG~\cite{lewis2020retrieval} uses dense retrieval over flat text, while GraphRAG~\cite{edge2024local} and LightRAG~\cite{guo2024lightrag} employ graph-based knowledge for more structured retrieval. Multimodal extensions such as WavRAG~\cite{chen2025wavrag} and VideoRAG~\cite{ren2025videorag} incorporate audio or video, enabling cross-modal fusion, but often lack explicit scene awareness. SceneRAG addresses this gap by combining scene-aware segmentation with graph-based, multimodal knowledge integration for more contextually grounded retrieval. Table~\ref{tab:rag_comparison} summarizes these representative RAG-based methods, highlighting key differences in modality, structure, and scene awareness.

\begin{table}[h]
\centering
\caption{Comparison of RAG-based Methods for Knowledge-Enhanced Generation.}
\label{tab:rag_comparison}
\begin{tabular}{lcccc}
\toprule
\textbf{Model} & \textbf{Modality} & \textbf{Knowledge Structure} & \textbf{Scene Awareness} & \textbf{Multimodal Fusion}  \\
\midrule
RAG           & Text              & Flat          & \xmark & \xmark  \\
GraphRAG      & Text              & Graph         & \xmark & \xmark  \\
LightRAG      & Text              & Graph         & \xmark & \xmark  \\
WavRAG        & Audio, Text       & Flat          & \xmark & \checkmark  \\
VideoRAG      & Video, Text       & Graph         & \xmark & \checkmark  \\
SceneRAG      & Video, Text       & Graph         & \checkmark & \checkmark  \\
\bottomrule
\end{tabular}
\end{table}

\paragraph{Scene Structuring for Retrieval and Reasoning.}
Segmenting videos into shots and scenes has long served as the foundation for video alignment and retrieval~\cite{yuan2007formal,pal2015video}. Traditional methods mainly detect shot transitions based on low-level visual features. Recent advances leverage self-supervised learning and contrastive objectives~\citep{wu2022scene,mun2022boundary,tan2024temporal} to improve boundary detection and capture richer scene structures, which benefit downstream retrieval tasks. In egocentric and cross-view scenarios, scene-level alignment is also key for connecting first- and third-person perspectives~\cite{xu2024retrieval}.However, most existing segmentation methods still lack semantic and narrative awareness, often resulting in boundaries that do not align with human-perceived scene changes. With the emergence of large pre-trained vision-language models~\citep{li2022blip,li2023videochat,zhang2023video}, research has begun to explore human-inspired, cognitively motivated scene segmentation—aiming to partition video into units that better reflect narrative flow and human perception.

\section{Method}
Section 3.1 outlines SceneRAG’s architecture; Section 3.2 introduces hybrid multimodal scene segmentation; Section 3.3 describes scene-based knowledge grounding via a lightweight graph; and Section 3.4 presents the retrieval-augmented generation module.

\subsection{Framework Overview}

Figure~1 shows the overall architecture of SceneRAG. Given a video $V$, we first divide it uniformly into fixed-length clips of duration $T$, resulting in a series of segments ${v_1, v_2, \dots, v_n}$, each with transcript snippets and timestamps $\mathcal{T} = {t_1, t_2, \dots, t_n}$. These multimodal inputs are fed into a large language model, $\mathrm{LLM}(\cdot)$, which produces an initial set of scene boundaries $\mathcal{S}^0$. The boundaries are then refined with heuristic rules to obtain the final scene set $\mathcal{S} = {S_1, S_2, \dots, S_m}$. As each new scene $S_i$ is identified, the knowledge graph $G = (N, E)$ is dynamically updated, where nodes $N$ denote scenes and entities, and edges $E$ represent relationships such as co-occurrence and temporal adjacency. Given a query $Q$, SceneRAG extracts keywords and retrieves candidate segments via multi-hop reasoning on $G$, aggregating context to generate a context-aware answer $A$.

\subsection{Automatic Scene Segmentation}

\paragraph{Chunk-wise Processing Pipeline.}
To facilitate scalable processing of long videos, we uniformly divide each video into a sequence of overlapping temporal chunks. Each chunk has a fixed length of $L = 5$ minutes with a $10$-second overlap between adjacent segments:
\begin{equation}
    \mathrm{Chunks} = \{\mathrm{Chunk}_k\}_{k=1}^K, \quad \mathrm{Chunk}_k \in [t_k^{\mathrm{start}}, t_k^{\mathrm{end}}]
    \tag{1}
\end{equation}
where $K$ denotes the total number of chunks for a given video.

\paragraph{Chunk-level Audio Transcription (ASR).}
For each chunk $\mathrm{Chunk}_k$, an automatic speech recognition system ~\cite{chunkssscformer,gandhi2023distil} is used to obtain a localized transcript $\mathcal{T}_k$:
\begin{equation}
    \mathcal{T}_k = \mathrm{ASR}(\mathrm{Chunk}_k)
    \tag{2}
\end{equation}

\paragraph{Scene Segmentation within Each Chunk.}
Given the chunk-wise transcript $\mathcal{T}_k$, we construct a prompt~\cite{edge2024local} $P_k$ and apply a language model $f$ to extract intra-chunk scene boundaries~\cite{ji2022learning}:
\begin{equation}
    \hat{\mathcal{S}}_k = f(P_k) = \{(t_{i,k}^{\mathrm{start}}, t_{i,k}^{\mathrm{end}}, \hat{s}_{i,k})\}_{i=1}^{N_k}
    \tag{3}
\end{equation}
where $t_{i,k}^{\mathrm{start}}$ and $t_{i,k}^{\mathrm{end}}$ represent the start and end timestamps of the $i$-th scene within chunk $k$, and $N_k$ is the total number of scenes predicted in that chunk.As in the global segmentation process, we enforce constraints on the number, length, and coverage of segments. If any constraints are violated, the system adaptively re-prompts the model and escalates to a stronger one if necessary.

\paragraph{Silence-Aware Refinement (per chunk).}
Silent intervals are extracted using the ASR system's capabilities, identifying segments where no speech is detected. These silent intervals $\mathcal{E}k$ provide important structural cues for refining segmentation. They help detect scene boundaries that may not be captured purely from textual signals. We classify each interval $e_{j,k}$ based on its duration $\Delta t_{j,k} = t_{j,k}^{\mathrm{end}} - t_{j,k}^{\mathrm{start}}$ and process it accordingly:

\begin{itemize}
  \item \textbf{Short Silence Assignment:} if $\Delta t_{j,k} \leq \epsilon$, set to 10 seconds by default, the silence is evenly split and assigned to adjacent scenes:

    \begin{equation}
    \text{Assign}(e_{j,k}) =
    \begin{cases}
    \text{Previous scene: } [t_{j,k}^{\mathrm{start}},\ t_{j,k}^{\mathrm{mid}}] \\
    \text{Next scene: } [t_{j,k}^{\mathrm{mid}},\ t_{j,k}^{\mathrm{end}}]
    \end{cases}, \quad t_{j,k}^{\mathrm{mid}} = \frac{t_{j,k}^{\mathrm{start}} + t_{j,k}^{\mathrm{end}}}{2}
    \tag{4}
    \end{equation}

    \item \textbf{Long Silence Promotion:} if $\Delta t_{j,k} >  \epsilon$, the silence is promoted as an independent scene:
    \begin{equation}
    S_{j,k}^{(\mathrm{silent})} = \left( t_{j,k}^{\mathrm{start}},\ t_{j,k}^{\mathrm{end}},\ \texttt{[SILENT]} \right)
    \tag{5}
    \end{equation}
\end{itemize}

These silence-derived segments are retained in downstream stages and later processed using multimodal models~\cite{yao2024minicpm} to extract scene-level representations. This prevents information loss from visually or acoustically rich transitions that may lack textual signals.

\paragraph{Temporal Adjustment and Post-hoc Correction.}
To ensure stable segmentation, we refine scene boundaries through two strategies. First, segments shorter than 10 seconds are merged with neighboring ones based on temporal and semantic proximity. Second, all scene boundaries are adjusted to align with sentence-level punctuation in the transcript $\mathcal{T}_k$. For each predicted scene $S_{i,k}$, we extract its aligned transcript segment:
\begin{equation}
    T_{i,k} = \mathcal{T}_k\big|_{[t_{i,k}^{\mathrm{start}},\ t_{i,k}^{\mathrm{end}}]}
    \tag{6}
\end{equation}

\paragraph{Final Output.}
The output of the segmentation module is a collection~\cite{sidiropoulos2011temporal} of aligned scene units that encapsulate temporal boundaries and transcripts:
\begin{equation}
    \mathcal{S} = \bigcup_{k=1}^K \left\{ \left( t_{i,k}^{\mathrm{start}},\ t_{i,k}^{\mathrm{end}},\\ T_{i,k} \right) \right\}_{i=1}^{N_k}
    \tag{7}
\end{equation}

This final set $\mathcal{S}$ provides a unified, structured representation of the video, serving as the foundation for downstream tasks such as summarization, classification, or multimodal grounding.

\subsection{Scene-based Multimodal Knowledge Grounding}

\paragraph{Multimodal Scene Representation.}
For each scene $S_j = (t_j^{\text{start}}, t_j^{\text{end}}, s_j)$, we leverage both the transcript $T_j$  and the visual appearance to construct a comprehensive and unified textual representation. To efficiently represent the visual content of each scene, we uniformly sample $k$ key frames (typically every 6 seconds, with $k \leq 10$), denoted as ${F_1, F_2, \ldots, F_k}$.  These sampled frames, together with the corresponding scene transcript $T_j$, are fed into a visual language model (VLM) to produce a natural language description $C_j$ that captures objects, actions, and high-level scene dynamics~\citep{peng2023multi,mithun2018learning,luo2020univl}:
\begin{equation}
    C_j = \mathrm{VLM}(T_j, \{F_1, \ldots, F_k\})
    \tag{8}
\end{equation}

At this stage, each scene has two complementary textual modalities: the transcript \( T_j \) (audio/text) and the visual description \( C_j \) (visual/text), enabling richer dual-channel knowledge extraction~\citep{xu2021videoclip}. To maximize cross-modal coverage, we extract entities and relations separately from \( C_j \) and \( T_j \) using large language models (LLMs)~\citep{edge2024local,guo2024lightrag}: \((N_j^{(\mathrm{vis})}, E_j^{(\mathrm{vis})})\) from \( C_j \), and \((N_j^{(\mathrm{asr})}, E_j^{(\mathrm{asr})})\) from \( T_j \). This dual-path strategy preserves modality-specific information and effectively alleviates the modality bias issues observed in prior work~\citep{ren2025videorag}.

\paragraph{Multimodal Entity and Relation Fusion.}
We merge entities and relations from both modalities into a unified scene-level knowledge set, using LLM-assisted disambiguation to align semantically equivalent entities and preserve cross-modal relations~\citep{lee2024multimodal,arnab2021unified}:
\begin{align}
    \tag{9}
    N_j &= \mathrm{Fuse}(N_j^{(\mathrm{vis})}, N_j^{(\mathrm{asr})}) \\
    \tag{10}
    E_j &= \mathrm{Fuse}(E_j^{(\mathrm{vis})}, E_j^{(\mathrm{asr})})
\end{align}
yielding a comprehensive set of entities $N_j$ and relations $E_j$ for each scene.

\paragraph{Knowledge Graph Construction and Integration.}
Adopting the incremental graph-building strategy from GraphRAG~\citep{edge2024local}, we assemble all scene-level entities and relations into a unified knowledge graph $G = (N, E)$:
\begin{align}
    G = (N, E) = \bigcup_{j=1}^{N'} (N_j, E_j)
    \tag{11}
\end{align}
where $N'$ is the total number of scenes. For multi-video corpora, we unify semantically equivalent entities and relations across videos for cross-video knowledge aggregation and entity enrichment~\citep{lee2024multimodal}. LLMs synthesize comprehensive entity descriptions from multiple scene contexts~\citep{peng2023multi}.

\paragraph{Multi-Modal Context Encoding}
Each scene’s visual description $C_j$ and transcript $T_j$ are concatenated into a unified text chunk $H_j$, allowing for richer contextual representation that captures both visual and textual cues. This chunk is then embedded using a pre-trained multimodal encoder~\cite{girdhar2023imagebind}.
\begin{align}
    H_j &= [C_j; T_j] \tag{12} \\
    e_t^j &= \mathrm{TEnc}(H_j) \tag{13}
\end{align}

where $C_j$ denotes the visual description of the scene and $T_j$ is its transcript. All scene embeddings are aggregated into a matrix $E_t \in \mathbb{R}^{N' \times d_t}$, where $N'$ is the number of scenes and $d_t$ is the embedding dimension. These embeddings are later utilized for efficient semantic retrieval and downstream integration with external knowledge sources.

\subsection{Scene-based Retrieval-Augmented Generation}

\paragraph{Token-Budgeted Scene Retrieval.}  
Given a query \(q\), we first encode it as \(e_q\) using the same text encoder as for scene embeddings~\cite{radford2021learning,girdhar2023imagebind}.  
The relevance between the query and each scene is measured by cosine similarity \(s_j=\cos(e_q,e_t^j)\), and we select a subset of scenes \(\mathcal{R}_q^*\) under a total token-length constraint \(\tau\):  
\[
  \mathcal{R}_q^* = \underset{\mathcal{R}\subseteq\{1,\dots,N'\}}{\arg\max}\sum_{j\in\mathcal{R}}s_j
  \quad\text{s.t.}\quad
  \sum_{j\in\mathcal{R}}\mathrm{Len}(C_j,T_j)\le\tau
  \tag{14}
\]
where $\text{Len}(C_j, T_j)$ is the total token length of the scene's content and $\tau$ is the maximum allowed token budget (e.g., 2400 tokens).

\begin{table*}[htbp]
\centering

\scriptsize
\caption{Comparison between SceneRAG and five baselines (NaiveRAG, GraphRAG\textsubscript{1} (GraphRAG-local), GraphRAG\textsubscript{2} (GraphRAG-global),  LightRAG and VideoRAG) on the LongerVideos dataset. Numbers indicate win-rates (\%).  The ``All'' column aggregates results from the three domains.}
\label{tab:win-rate}
\begin{tabular}{l
  >{\centering\arraybackslash}p{1.0cm}
  >{\centering\arraybackslash}p{1.0cm}
  >{\centering\arraybackslash}p{1.0cm}
  >{\centering\arraybackslash}p{1.0cm}
  >{\centering\arraybackslash}p{1.0cm}
  >{\centering\arraybackslash}p{1.0cm}
  >{\centering\arraybackslash}p{1.0cm}
  >{\centering\arraybackslash}p{1.0cm}
}
\toprule
 & \multicolumn{2}{c}{Lecture} & \multicolumn{2}{c}{Documentary} & \multicolumn{2}{c}{Entertainment} & \multicolumn{2}{c}{All} \\
\cmidrule(lr){2-3} \cmidrule(lr){4-5} \cmidrule(lr){6-7} \cmidrule(lr){8-9}

 & NaiveRAG & \textbf{SceneRAG} & NaiveRAG & \textbf{SceneRAG} & NaiveRAG & \textbf{SceneRAG} & NaiveRAG & \textbf{SceneRAG} \\
\midrule
\raggedright Comprehensiveness & 35.9\% & \textbf{64.1\%} & 30.7\% & \textbf{69.3\%} & 35.1\% & \textbf{64.9\%} & 34.8\% & \textbf{65.2\%} \\
\raggedright Empowerment       & 34.2\% & \textbf{65.8\%} & 26.8\% & \textbf{73.2\%} & 33.9\% & \textbf{66.1\%} & 32.8\% & \textbf{67.2\%} \\
\raggedright Trustworthiness   & 37.0\% & \textbf{63.0\%} & 28.2\% & \textbf{71.8\%} & 37.5\% & \textbf{62.5\%} & 35.4\% & \textbf{64.6\%} \\
\raggedright Depth             & 35.3\% & \textbf{64.7\%} & 27.3\% & \textbf{72.7\%} & 34.2\% & \textbf{65.8\%} & 33.6\% & \textbf{66.4\%} \\
\raggedright Density           & 50.3\% & \textbf{49.7\%} & 50.4\% & \textbf{49.6\%} & 48.4\% & \textbf{51.6\%} & 50.0\% & \textbf{50.0\%} \\
\raggedright Overall Winner    & 36.2\% & \textbf{63.8\%} & 28.7\% & \textbf{71.3\%} & 34.5\% & \textbf{65.5\%} & 34.5\% & \textbf{65.5\%} \\

\midrule
 & GraphRAG\textsubscript{1} & \textbf{SceneRAG} & GraphRAG\textsubscript{1} & \textbf{SceneRAG} & GraphRAG\textsubscript{1} & \textbf{SceneRAG} & GraphRAG\textsubscript{1} & \textbf{SceneRAG} \\
\midrule
\raggedright Comprehensiveness & 32.9\% & \textbf{67.1\%} & 37.7\% & \textbf{62.3\%} & 39.4\% & \textbf{60.6\%} & 35.0\% & \textbf{65.0\%} \\
\raggedright Empowerment       & 28.5\% & \textbf{71.5\%} & 33.6\% & \textbf{66.4\%} & 34.8\% & \textbf{65.2\%} & 30.7\% & \textbf{69.3\%} \\
\raggedright Trustworthiness   & 31.6\% & \textbf{68.4\%} & 32.9\% & \textbf{67.1\%} & 35.3\% & \textbf{64.7\%} & 32.6\% & \textbf{67.4\%} \\
\raggedright Depth             & 29.2\% & \textbf{70.8\%} & 33.6\% & \textbf{66.4\%} & 33.6\% & \textbf{66.4\%} & 30.9\% & \textbf{69.1\%} \\
\raggedright Density           & 36.3\% & \textbf{63.7\%} & 45.8\% & \textbf{54.2\%} & 42.0\% & \textbf{58.0\%} & 39.1\% & \textbf{60.9\%} \\
\raggedright Overall Winner    & 29.5\% & \textbf{70.5\%} & 35.1\% & \textbf{64.9\%} & 35.0\% & \textbf{65.0\%} & 31.6\% & \textbf{68.4\%} \\

\midrule
 & GraphRAG\textsubscript{2} & \textbf{SceneRAG} & GraphRAG\textsubscript{2} & \textbf{SceneRAG} & GraphRAG\textsubscript{2} & \textbf{SceneRAG} & GraphRAG\textsubscript{2} & \textbf{SceneRAG} \\
\midrule
\raggedright Comprehensiveness & 29.9\% & \textbf{70.1\%} & 35.7\% & \textbf{64.3\%} & 40.0\% & \textbf{60.0\%} & 32.9\% & \textbf{67.1\%} \\
\raggedright Empowerment       & 26.1\% & \textbf{73.9\%} & 32.5\% & \textbf{67.5\%} & 37.5\% & \textbf{62.5\%} & 29.4\% & \textbf{70.6\%} \\
\raggedright Trustworthiness   & 26.2\% & \textbf{73.8\%} & 28.1\% & \textbf{71.9\%} & 31.4\% & \textbf{68.6\%} & 27.5\% & \textbf{72.5\%} \\
\raggedright Depth             & 25.8\% & \textbf{74.2\%} & 30.1\% & \textbf{69.9\%} & 34.8\% & \textbf{65.2\%} & 28.3\% & \textbf{71.7\%} \\
\raggedright Density           & 35.5\% & \textbf{64.5\%} & 51.3\% & \textbf{48.7\%} & 49.9\% & \textbf{50.1\%} & 41.2\% & \textbf{58.8\%} \\
\raggedright Overall Winner    & 26.0\% & \textbf{74.0\%} & 32.7\% & \textbf{67.3\%} & 36.2\% & \textbf{63.8\%} & 29.2\% & \textbf{70.8\%} \\

\midrule
 & LightRAG & \textbf{SceneRAG} & LightRAG & \textbf{SceneRAG} & LightRAG & \textbf{SceneRAG} & LightRAG & \textbf{SceneRAG} \\
\midrule
\raggedright Comprehensiveness & 32.4\% & \textbf{67.6\%} & 31.9\% & \textbf{68.1\%} & 33.5\% & \textbf{66.5\%} & 32.5\% & \textbf{67.5\%} \\
\raggedright Empowerment       & 28.5\% & \textbf{71.5\%} & 30.1\% & \textbf{69.9\%} & 30.4\% & \textbf{69.6\%} & 29.2\% & \textbf{70.8\%} \\
\raggedright Trustworthiness   & 31.5\% & \textbf{68.5\%} & 28.7\% & \textbf{71.3\%} & 31.8\% & \textbf{68.2\%} & 31.0\% & \textbf{69.0\%} \\
\raggedright Depth             & 28.7\% & \textbf{71.3\%} & 27.8\% & \textbf{72.2\%} & 29.6\% & \textbf{70.4\%} & 28.7\% & \textbf{71.3\%} \\
\raggedright Density           & 42.6\% & \textbf{57.4\%} & 50.3\% & \textbf{49.7\%} & 42.9\% & \textbf{57.1\%} & 44.1\% & \textbf{55.9\%} \\
\raggedright Overall Winner    & 29.8\% & \textbf{70.2\%} & 29.5\% & \textbf{70.5\%} & 31.1\% & \textbf{68.9\%} & 30.0\% & \textbf{70.0\%} \\

\midrule
& VideoRAG & \textbf{SceneRAG} & VideoRAG & \textbf{SceneRAG} & VideoRAG & \textbf{SceneRAG} & VideoRAG & \textbf{SceneRAG} \\
\midrule
\raggedright Comprehensiveness & 43.8\% & \textbf{56.2\%} & 40.2\% & \textbf{59.8\%} & 43.3\% & \textbf{56.7\%} & 43.1\% & \textbf{56.9\%} \\
\raggedright Empowerment       & 42.6\% & \textbf{57.4\%} & 39.0\% & \textbf{61.0\%} & 42.4\% & \textbf{57.6\%} & 41.9\% & \textbf{58.1\%} \\
\raggedright Trustworthiness   & 42.4\% & \textbf{57.6\%} & 37.4\% & \textbf{62.6\%} & 40.0\% & \textbf{60.0\%} & 41.0\% & \textbf{59.0\%} \\
\raggedright Depth             & 42.5\% & \textbf{57.5\%} & 38.3\% & \textbf{61.7\%} & 42.3\% & \textbf{57.7\%} & 41.7\% & \textbf{58.3\%} \\
\raggedright Density           & 47.7\% & \textbf{52.3\%} & 47.2\% & \textbf{52.8\%} & 50.9\% & \textbf{49.1\%} & 48.2\% & \textbf{51.8\%} \\
\raggedright Overall Winner    & 42.8\% & \textbf{57.2\%} & 38.9\% & \textbf{61.1\%} & 42.7\% & \textbf{57.3\%} & 42.0\% & \textbf{58.0\%} \\
\bottomrule
\end{tabular}
\end{table*}
\paragraph{Scene-aware Answer Generation.}
Given a query $q$, we first extract salient keywords using an LLM~\cite{guo2024lightrag}, and generate query-focused visual captions for each retrieved scene by prompting a VLM with these keywords, the transcript, and sampled frames. We then aggregate all focused captions, transcripts, and knowledge-graph components from the relevant scenes to construct the context:
\begin{equation}
  C_q = \mathrm{Concat}\bigl(\{\hat{C}_j,\,T_j,\,(N_j,E_j)\mid j\in\mathcal{R}_q^*\}\bigr)
  \tag{15}
\end{equation}
where $\hat{C}_j$ is the query-focused visual caption for scene $j$, $T_j$ is the transcript, $(N_j, E_j)$ are the node and edge sets of the knowledge graph for $j$, $\mathcal{R}_q^*$ is the set of relevant scenes retrieved for $q$, and $C_q$ is the constructed context. The answer is then generated by feeding the query and context into an LLM:
\begin{equation}
  A_q = \mathrm{LLM}(q,\,C_q)
  \tag{16}
\end{equation}
where $A_q$ denotes the generated answer for the input query $q$, using context $C_q$.

\section{Experiments}

\subsection{Experimental Setup}
\paragraph{Dataset}
We evaluate our approach on the \textsc{LongerVideos}~\cite{ren2025videorag} benchmark—a large-scale dataset curated to assess models on extended video content. It contains 164 videos across 22 lists, totaling over 134 hours, and spans three challenging domains: lectures, documentaries, and entertainment. Compared to prior benchmarks focused on short clips or single-topic narratives, \textsc{LongerVideos} offers a more realistic and demanding testbed. Its videos often exceed 30 minutes, exhibit multiple semantic shifts, and contain rich multimodal signals, making it well-suited for evaluating scene-level reasoning, retrieval-augmented generation, and temporal comprehension under complex structure.

\paragraph{Baseline Methods.}
To comprehensively evaluate the effectiveness of our proposed method, we compare SceneRAG with a range of baselines across three categories: (1) Text-centric RAG methods, including NaiveRAG~\cite{gao2023retrieval}, GraphRAG (both local and global variants)\cite{edge2024local}, and LightRAG\cite{guo2024lightrag}, which operate solely on text transcripts without modeling multimodal or temporal structure; (2) Video-centric RAG frameworks, including VideoRAG~\cite{ren2025videorag}, which builds a large-scale multimodal graph across videos, performing semantic retrieval based on visual and textual features; (3) General long video understanding models, such as LLaMA-VID~\cite{li2024llama}, VideoAgent~\cite{fan2024videoagent}, and NotebookLM, designed for long-context vision-language understanding but not following the RAG paradigm.

\paragraph{Evaluation Protocol.}
We adopt a two-part LLM-based evaluation protocol for long-context video understanding: \textit{Win-Rate Comparison} and \textit{Quantitative Scoring}. Both use five human-centric dimensions: (1) \textit{Comprehensiveness}—coverage of query intent; (2) \textit{Empowerment}—perceived helpfulness; (3) \textit{Trustworthiness}—factual accuracy; (4) \textit{Depth}—reasoning quality; and (5) \textit{Density}—amount of relevant, non-redundant information. Scores are averaged across three domains (lectures, documentaries, entertainment). Win-rate is based on LLM-judged pairwise comparisons; quantitative scores are LLM-rated against NaiveRAG on a 5-point Likert scale.

\paragraph{Implementation Details.}
We adopt the same experimental setup as VideoRAG, ensuring fair comparisons across models. All methods use GPT-4o-mini for generation, with shared prompt formatting and input encoding pipelines. ASR transcripts are generated using Distil-Whisper~\cite{radford2023robust,gandhi2023distil}. Visual and textual features are extracted using ImageBind~\cite{girdhar2023imagebind}, which serves as the unified multimodal encoder \( \text{MEnc}(\cdot) \). All baselines utilize grounded textual knowledge, including visual captions and transcripts from videos, and apply the same chunk-splitting protocol as our method. Entity- and chunk-level retrieval rely on OpenAI’s \texttt{text-embedding-3-small} model, while vision-language modeling uses MiniCPM-V~\cite{yao2024minicpm}. Videos are sampled uniformly at 6-second intervals, and transcript, frame, metadata inputs, as well as all retrieved segments, are processed through a shared normalization pipeline for consistency.

\subsection{Main Results}

We evaluate SceneRAG on the \textit{LongerVideos} dataset using a two-part win-rate study, employing three language models (\textbf{GPT-4o-mini}, \textbf{GPT-4.1-mini}, and \textbf{GPT-4.1-nano}) to ensure robustness and minimize model bias. Results are aggregated across models unless otherwise specified, with per-model details in Appendix~C.

\paragraph{Comparison with Text-Based RAG Baselines.}
We compare SceneRAG against five strong text-centric retrieval-augmented generation baselines—NaiveRAG, GraphRAG-local, GraphRAG-global, LightRAG, and VideoRAG—on the LongerVideos dataset. Results in Table~\ref{tab:win-rate} demonstrate that SceneRAG consistently outperforms all baselines across domains and evaluation dimensions. SceneRAG achieves the highest average overall win-rate of 70.8\% when compared to GraphRAG-global, marking it as the strongest model in the comparison. Gains are particularly substantial in dimensions such as Empowerment (up to 73.9\%) and Depth (up to 74.2\%), underscoring SceneRAG’s ability to integrate temporally dispersed multimodal evidence into coherent, in-depth scene-level outputs. These consistent improvements suggest that scene-aware structuring and multimodal graph-based retrieval provide a significant advantage in understanding long-form videos.

\paragraph{Comparison with Multimodal RAG Baseline.}
We further compare SceneRAG with VideoRAG, a representative multimodal RAG system that retrieves over concatenated video-text representations. SceneRAG outperforms VideoRAG across all five human evaluation metrics, achieving an overall win rate of 56.9\% versus 43.1\%. While VideoRAG relies on dense video-text embeddings without explicit structural modeling, SceneRAG incorporates scene-aware segmentation and graph-based propagation to improve contextual alignment. This leads to more semantically coherent and temporally grounded outputs, especially in scenarios requiring fine-grained disambiguation or multi-hop reasoning over long-form content.Compared to VideoRAG, which relies on fixed-interval segmentation, our approach aligns better with semantic boundaries, leading to more coherent and relevant outputs.

\begin{figure}[h]
    \centering
    \includegraphics[width=1\textwidth]{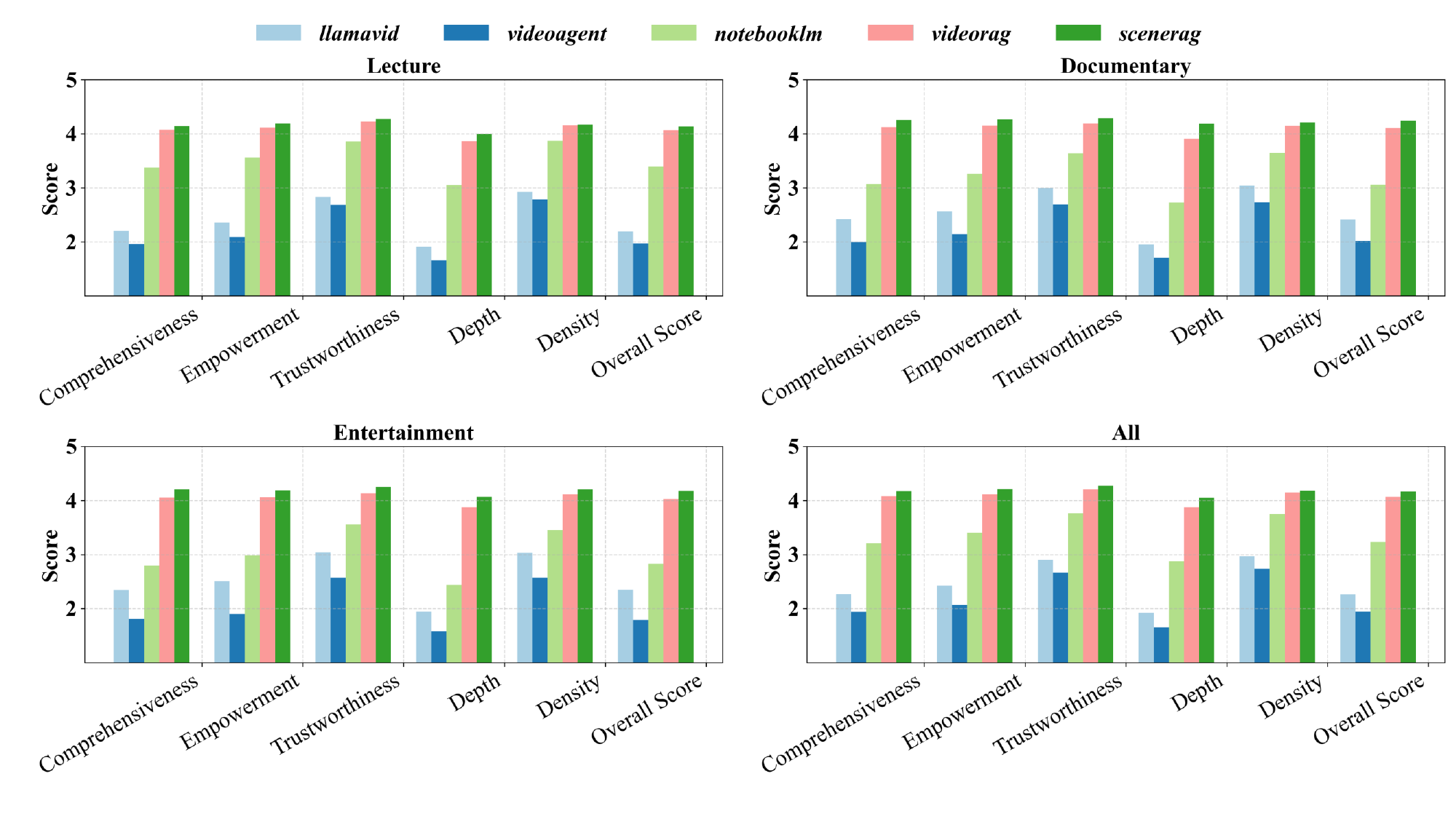} 
    \caption{Quantitative Comparison with NaiveRAG Using a 5-Point Likert Scale. The ``All'' column aggregates results from the three domains.}
    \label{fig:quancompan}
\end{figure}
\paragraph{Comparison with Large Vision-Language Models.}
As shown in Figure~\ref{fig:quancompan}, SceneRAG consistently outperforms large vision-language models (LVMs) across all four domains, with particularly strong gains in lecture and documentary content that require long-range, multimodal reasoning. This advantage stems from SceneRAG’s dynamic scene segmentation and graph-based retrieval. While LVMs typically process fixed-length video chunks without structural cues, SceneRAG organizes content into semantically coherent scenes, enabling better alignment of audio, visual, and textual information. The graph module further supports cross-scene linking and retrieval, enhancing contextual understanding in complex narratives.These design choices allow SceneRAG to better handle temporal dependencies and topic transitions, especially in genres where meaning builds progressively over time.

\subsection{Case Study}

SceneRAG’s ability to locate and leverage cohesive segments in long-form videos is evaluated via a case study on the “Is This the End of RAG? Anthropic’s NEW Prompt Caching” lecture. For the query “How does prompt caching compare to traditional RAG in cost and efficiency?”, SceneRAG identifies non-contiguous, complementary segments—[23.96–66.08 s] and [213.12–292.48 s]—totaling 122 s. These clips provide a complete, focused answer despite being distributed across the timeline.

\paragraph{Coherent Segmentation} Traditional methods relying on dense chunking or uniform sliding windows often fragment context, leading to partial or diluted answers. In contrast, SceneRAG’s scene-level segmentation ensures that each retrieved segment preserves a coherent narrative unit. The first retrieved scene introduces the motivation behind prompt caching and frames the cost/latency benefits in contrast to traditional RAG approaches. The second scene dives into concrete quantitative comparisons (e.g., 80–90\% reduction in latency and cost), clarifies trade-offs like cache write costs, and juxtaposes Anthropic’s and Gemini's implementations.

\begin{table}[h]
\centering
\caption{Segment Retrieval Comparison. Symbols denote retrieval quality: \checkmark\ indicates all retrieved segments are strongly coherent, relevant, or fully answer the query with key points; \xmark\ means some segments are fragmented, loosely related, or partially address key information.}

\begin{tabular}{lccccc}
\toprule
\textbf{Model} & \textbf{\#Segments} & \textbf{Duration (s)} & \textbf{Coherence} & \textbf{Relevance} & \textbf{Key Points} \\
\midrule
SceneRAG & 2 & 122 & \checkmark & \checkmark & \checkmark \\
VideoRAG & 9 & 270 & \xmark & \xmark & \checkmark \\
\bottomrule
\end{tabular}
\label{tab:retrieval_comparison}
\end{table}

\paragraph{Precise Retrieval} SceneRAG retrieves only the most relevant, semantically cohesive segments, ensuring responses remain focused. By avoiding irrelevant content, SceneRAG reduces processing time. For example, it retrieved two highly relevant segments totaling 122 seconds. This precision improves efficiency and minimizes distractions. In contrast, VideoRAG retrieved nine segments (270 seconds), including less relevant content, increasing processing time and task complexity. Such over-retrieval is especially problematic for large videos or when VLM resources are limited. The impact is shown in Table~\ref{tab:retrieval_comparison}.

\paragraph{Domain Adaptability} SceneRAG’s alignment of retrieval granularity with natural discourse boundaries proves especially beneficial in structured content such as technical lectures. Its ability to aggregate temporally distant yet topically connected segments enables accurate, context-rich responses, even when critical information is spread across multiple video parts. This domain adaptability makes SceneRAG an optimal choice for queries spanning multiple scenes in long-form content.

\subsection{Ablation}

\paragraph{Ablation Study.} 
We perform an ablation study to evaluate the impact of components in our scene segmentation pipeline. As shown in Table~\ref{tab:ablation}, LLM-based scene-aware segmentation enhances performance across all metrics compared to raw input. Additional rule-based refinements—Short segment allocation and mute processing—yield further improvements, particularly in Depth and Comprehensiveness. This underscores the importance of accurate scene segmentation for enhancing retrieval relevance and generation quality in long-context video understanding.

\begin{table}[H]
\caption{Results of ablation study. ``/'' denotes fixed segmentation without LLM assistance; ``+LLM'' adds LLM-guided segmentation; ``+LLM+Rules'' further incorporates rule-based refinement.}
\label{tab:ablation}
\centering

\resizebox{\textwidth}{!}{
\begin{tabular}{
    l
    |ccc
    |ccc
    |ccc
    |ccc
    }
\toprule
& 
\multicolumn{3}{c|}{\textbf{\large Lecture}} &
\multicolumn{3}{c|}{\textbf{\large Documentary}} &
\multicolumn{3}{c|}{\textbf{\large Entertainment}} &
\multicolumn{3}{c}{\textbf{\large All}} \\
\cmidrule(lr){2-4}\cmidrule(lr){5-7}\cmidrule(lr){8-10}\cmidrule(lr){11-13}
& 
\textbf{\large /} & \textbf{\large +LLM} & \makecell{\textbf{\large +LLM}\\\textbf{\large +Rules}} &
\textbf{\large /} & \textbf{\large +LLM} & \makecell{\textbf{\large +LLM}\\\textbf{\large +Rules}} &
\textbf{\large /} & \textbf{\large +LLM} & \makecell{\textbf{\large +LLM}\\\textbf{\large +Rules}} &
\textbf{\large /} & \textbf{\large +LLM} & \makecell{\textbf{\large +LLM}\\\textbf{\large +Rules}} \\
\midrule
Comprehensiveness & 4.35 & 4.36 & 4.51 & 4.33 & 4.44 & 4.56 & 4.30 & 4.39 & 4.52 & 4.35 & 4.38 & 4.52 \\
Empowerment       & 4.49 & 4.45 & 4.57 & 4.38 & 4.56 & 4.64 & 4.29 & 4.42 & 4.56 & 4.43 & 4.47 & 4.58 \\
Trustworthiness   & 4.44 & 4.46 & 4.55 & 4.39 & 4.43 & 4.61 & 4.37 & 4.53 & 4.58 & 4.42 & 4.47 & 4.57 \\
Depth             & 4.22 & 4.22 & 4.44 & 4.10 & 4.31 & 4.56 & 4.03 & 4.29 & 4.45 & 4.16 & 4.25 & 4.47 \\
Density           & 4.51 & 4.53 & 4.60 & 4.53 & 4.68 & 4.62 & 4.46 & 4.46 & 4.62 & 4.50 & 4.55 & 4.61 \\
Overall Score     & 4.37 & 4.35 & 4.49 & 4.31 & 4.40 & 4.54 & 4.24 & 4.38 & 4.51 & 4.33 & 4.37 & 4.50 \\
\bottomrule
\end{tabular}
}
\end{table}

\paragraph{Effect of Scene-Based Segmentation.}
Compared to fixed-length segmentation, scene-based segmentation aligns boundaries with actual semantic shifts, preserving contextual coherence and narrative flow. As shown in Table~\ref{tab:nodes-edges}, this approach results in denser knowledge graphs~\cite{zhang2025rakg}, capturing more entities and relations. By structuring information around scenes, relevant entities are more tightly connected, facilitating multi-hop retrieval and improving both retrieval precision and answer generation.

\begin{table}[H]
  \centering
  \caption{Graph Expansion Comparison. Values are normalized to the fixed-segmentation baseline.}

  \label{tab:nodes-edges}
  \begin{tabular}{lccc}
    \toprule
           &\raisebox{-0.5\height} /    & \raisebox{-0.5\height}{+LLM}  & \raisebox{-0.5\height}{\shortstack{+LLM\\+Rules}} \\
    \midrule
    nodes  & 1    & 1.13           & 1.29                                \\
    edges  & 1    & 1.22           & 1.34                                \\
    \bottomrule
  \end{tabular}
\end{table}

\section{Conclusion}
Long video analysis poses significant challenges in computer vision and machine learning, as existing models struggle to capture complex temporal dependencies and dynamic transitions across lengthy sequences. Inspired by human perception, our model, SceneRAG, addresses these challenges by segmenting video content into evolving, semantically coherent scenes, enhancing understanding of long-term dependencies and context shifts. By focusing on scene-level granularity, SceneRAG efficiently tracks and predicts scene boundaries and transitions, while preserving narrative continuity. In the future, we plan to incorporate advanced spatiotemporal cues and multi-modal data to further improve scene segmentation, enabling more accurate scene prediction, dynamic scene understanding, and flexible handling of diverse video content.

\bibliography{Reference}
\normalsize
\clearpage  


\appendix

\section{Implementation Details}
SceneRAG adopts a modular architecture inspired by the experimental setup of VideoRAG. Input videos are segmented into overlapping 5-minute chunks using \texttt{FFmpeg}, with a 10-second overlap to preserve contextual continuity. We employ \texttt{faster-distil-whisper-large-v3} for automatic speech recognition (ASR), which provides timestamped transcripts and identifies low-energy silence intervals.

Scene segmentation is performed via an LLM-based strategy, as shown in Algorithm~\ref{alg:llm_segmentation}. The system starts with a lightweight model GPT-4o-mini) using temperature = 0.7 and top\_p = 0.95. We utilize the prompts illustrated in figure~\ref{figscene-segmentation}to guide the model's output. If the returned scene boundaries are inconsistent or violate duration constraints, we iteratively retry up to four times, potentially escalating to a more powerful model (GPT-4o) and adapting the prompts based on the type of failure, such as overly long or short segments.

The output is further refined through post-processing heuristics. Silence-aware adjustment and interpolation are critical, as silent segments are typically omitted from ASR transcripts and manifest as temporal gaps between dialogue-based scene predictions. We detect low-energy silence intervals to inform scene boundaries, promoting long silences (e.g., greater than 10 seconds) as new segment breaks, while redistributing shorter silences across neighboring scenes or filling them through boundary expansion to ensure continuous temporal coverage. Additionally, we merge scene segments that are shorter than 10 seconds with adjacent segments. This merging strategy takes into account both segment duration and textual coherence, favoring merge directions that result in a smoother scene flow.
To enhance the accuracy of scene segmentation, we implement a scene correction mechanism, utilizing prompts depicted in figure~\ref{fig:scene-segmentation-repair}to improve the model's understanding and judgment of scene boundaries. For visual understanding, we extract representative video frames and encode them using MiniCPM-V-2\_6-int4. To unify information across video, audio, and text, we employ ImageBind for cross-modal embedding into a shared vector space.
The system infrastructure comprises a KV store for caching video paths, transcript blocks, and LLM results, a graph database to represent inter-scene relations and entities, and a vector database for multimodal dense retrieval. All experiments are executed on a single NVIDIA RTX 3090 GPU.

\begin{algorithm}[H]
\caption{LLM-Based Video Scene Segmentation}
\label{alg:llm_segmentation}
\begin{algorithmic}[1]
\REQUIRE Transcript with timestamps $\mathcal{T}$, Global config $\mathcal{C}$, Video duration $D$
\ENSURE Scene time intervals $\{t_1, t_2, \dots, t_n\}$, Scene descriptions $\{s_1, s_2, \dots, s_n\}$
\STATE Format segmentation prompt $P$ using $\mathcal{T}$
\STATE Initialize retry count $r \gets 0$
\REPEAT
    \STATE $R \gets \mathrm{LLMFunc}(P,\, \text{history})$
    \STATE $T \gets \mathrm{ExtractTimeRanges}(R)$
    \IF{$\mathrm{CheckTimeRanges}(T, D)$ is valid}
        \STATE \textbf{break}
    \ELSE
        \STATE $P' \gets \mathrm{ChooseFixPrompt}(\text{error})$
        \STATE $P \gets P'$
        \STATE $r \gets r + 1$
    \ENDIF
\UNTIL{$r \geq 4$}
\IF{no valid $S$ is found}
    \STATE \textbf{Return} default time intervals and empty description list
\ENDIF
\STATE $S \gets \mathrm{SplitText}(R)$
\STATE $T \gets \mathrm{FillTimeGaps}(S, T, D)$
\STATE $T, S \gets \mathrm{MergeIntervals}(T, S)$
\STATE \textbf{Return} $(T, S)$
\end{algorithmic}
\end{algorithm}

\begin{figure}
    \centering
    \begin{tcolorbox}[
        colback=green!5, 
        colframe=green!80!black, 
        title={Scene\_Segmentation\_repaired},
        sharp corners=southwest, 
        rounded corners=northeast, 
        boxrule=0.8mm 
    ]
    
    \textbf{scene\_segmentation\_too\_little} \\
    Output Error Correction Request: \\
    Too few time ranges. Need at least 3 segments. The previous output has errors. Please verify and correct the following:

    1. Ensure each scene has a duration between 15 and 60 seconds.

    2. Verify that the scenes are divided correctly based on the content.

    3. Ensure each scene starts with a time mark.

    4. Ensure each scene contains detailed descriptions, dialogues, or events to form a coherent narrative unit.

    Please maintain the required format in your response.

    \vspace{1em}

    \textbf{scene\_segmentation\_too\_short} \\
    Output Error Correction Request: \\
    Some scenes have been split with time ranges that are too short. Merge scenes that are too short to meet the minimum duration requirement. Please verify and correct the following:

    1. Ensure each scene has a duration between 15 and 60 seconds.

    2. Verify that the scenes are divided correctly based on the content.

    3. Ensure each scene starts with a time mark.

    4. Ensure each scene contains detailed descriptions, dialogues, or events to form a coherent narrative unit.

    5. Each scene should follow the previous one in a logical time sequence without gaps or overlaps.

    Please maintain the required format in your response.

    \vspace{1em}

    \textbf{scene\_segmentation\_too\_long} \\
    Output Error Correction Request: \\
    Some scenes have been split with time ranges that are too long. Split scenes that exceed the maximum duration into smaller segments. Please verify and correct the following:

    1. The duration of each scene should ideally not exceed 60 seconds.

    2. Verify that the scenes are divided correctly based on the content.

    3. Ensure each scene starts with a time mark.

    4. Ensure each scene contains detailed descriptions, dialogues, or events to form a coherent narrative unit.

    5. Each scene should follow the previous one in a logical time sequence without gaps or overlaps.

    Please maintain the required format in your response.

    \end{tcolorbox}
    \caption{Error correction instructions for scene segmentation.}
    \label{fig:scene-segmentation-repair}
\end{figure}

\begin{figure}
    \centering
   \begin{tcolorbox}[
    colback=green!5, 
    colframe=green!80!black, 
    title={Scene\_Segmentation},
    sharp corners=southwest, 
    rounded corners=northeast, 
    boxrule=0.8mm, 
]

\textbf{-Goal-} \\
The task is to segment the input text into distinct scenes based on the given criteria. The segmentation should be done purely based on the content provided, without the need for summarization or interpretation.

\textbf{-Steps-}

1. \textbf{Scene Identification and Segmentation} \\
- Identify distinct scenes in the text, and need to reflect on why these scenes are segmented. The segmentation should be based solely on the content and structure of the text. \\
- Ensure each scene contains detailed descriptions, dialogues, or events to form a coherent narrative unit, and must not consist of a single sentence.

2. \textbf{Time Range and Scene Delimiters} \\
- For each scene, record the time range (if available) at the beginning in the format \texttt{[start\_time -> end\_time]}. \\
- Each scene should follow the previous one in a logical time sequence without gaps or overlaps. \\
- Add the scene content after the time range. \\
- The duration of each scene is between 15 and 60 seconds, except for those that you think are special. \\
- End each scene with \texttt{\{record\_delimiter\}} (except the last scene).

3. \textbf{Final Marker} \\
- After all scenes, add \texttt{\{completion\_delimiter\}} to indicate the end of the task.

4. \textbf{Output Format} \\
- Return the segmented text as a list of scenes. \\
- Output format Example: \\
Scene 1\texttt{\{record\_delimiter\}} \\
Scene 2\texttt{\{record\_delimiter\}} \\
Scene 3\texttt{\{record\_delimiter\}} \\
Scene 4\texttt{\{completion\_delimiter\}}

- Output only the segmented text without additional explanations.
\#\#\#\#\#\#\#\#\#\#\#\#\#\#\#\#\#\#\#\#\#\#
Text: \texttt{\{input\_text\}}

\end{tcolorbox}
    \caption{Instructions for scene segmentation and formatting.}
    \label{figscene-segmentation}
\end{figure}

\section{Limitations and Future Work}
SceneRAG has two main limitations. First, the scene segmentation pipeline relies heavily on timestamped transcripts from ASR and large language models guided by hand-crafted prompts. This introduces sensitivity to transcription errors and prompt formulation, which can lead to suboptimal or inconsistent scene boundaries—particularly in noisy audio conditions or ambiguous dialogue contexts. Second, the current pipeline underutilizes non-verbal signals: while visual embeddings are incorporated downstream, the segmentation process is primarily text-driven. As a result, visually grounded transitions—such as emotional shifts, scene composition changes, or camera cuts without dialogue—may go undetected or misaligned.To address these challenges, we envision two future directions. One is to incorporate low-level visual and audio cues (e.g., shot boundary detection, background music changes, facial expression shifts) into the segmentation process to capture non-verbal scene transitions. The other is to reduce reliance on handcrafted prompts by leveraging prompt-free or instruction-tuned models that can better generalize across content domains.

\section{Supplementary Results} 
To ensure the robustness and fairness of our win-rate evaluation, we employ three language models—GPT-4o-mini, GPT-4.1-mini, and GPT-4.1-nano—to independently assess the outputs. For each comparison instance, we conduct two separate evaluations by swapping the order of the candidate answers, thereby mitigating potential position bias. The win-rate results reported for each model are averaged over both answer orders. Detailed per-model win-rate statistics are presented in the accompanying tables/figures. Notably, SceneRAG achieves even better performance when evaluated with the latest GPT-4.1-mini model, consistently outperforming baseline methods across all evaluation metrics and domains. This further demonstrates the robustness and effectiveness of SceneRAG, particularly when assessed by advanced open-domain language models.

\begin{table*}
\centering
\scriptsize
\caption{Comparison of Winning Rates by GPT-4o-mini.}
\begin{tabular}{l
  >{\centering\arraybackslash}p{1.0cm}
  >{\centering\arraybackslash}p{1.0cm}
  >{\centering\arraybackslash}p{1.0cm}
  >{\centering\arraybackslash}p{1.0cm}
  >{\centering\arraybackslash}p{1.0cm}
  >{\centering\arraybackslash}p{1.0cm}
  >{\centering\arraybackslash}p{1.0cm}
  >{\centering\arraybackslash}p{1.0cm}
}
\toprule
 & \multicolumn{2}{c}{Lecture} & \multicolumn{2}{c}{Documentary} & \multicolumn{2}{c}{Entertainment} & \multicolumn{2}{c}{All} \\
\cmidrule(lr){2-3} \cmidrule(lr){4-5} \cmidrule(lr){6-7} \cmidrule(lr){8-9}

 & NaiveRAG & SceneRAG & NaiveRAG & SceneRAG & NaiveRAG & SceneRAG & NaiveRAG & SceneRAG \\
\midrule
\raggedright Comprehensiveness & 42.2\% & \textbf{57.8\%} & 36.4\% & \textbf{63.6\%} & 37.5\% & \textbf{62.5\%} & 40.2\% & \textbf{59.8\%} \\
\raggedright Empowerment       & 39.8\% & \textbf{60.2\%} & 30.3\% & \textbf{69.7\%} & 35.7\% & \textbf{64.3\%} & 37.2\% & \textbf{62.8\%} \\
\raggedright Trustworthiness   & 39.8\% & \textbf{60.2\%} & 31.1\% & \textbf{68.9\%} & 35.7\% & \textbf{64.3\%} & 37.4\% & \textbf{62.6\%} \\
\raggedright Depth             & 39.8\% & \textbf{60.2\%} & 33.3\% & \textbf{66.7\%} & 37.1\% & \textbf{62.9\%} & 38.0\% & \textbf{62.0\%} \\
\raggedright Density           & 40.3\% & \textbf{59.7\%} & 37.3\% & \textbf{62.7\%} & 37.1\% & \textbf{62.9\%} & 39.1\% & \textbf{60.9\%} \\
\raggedright Overall Winner    & 42.0\% & \textbf{58.0\%} & 35.5\% & \textbf{64.5\%} & 37.1\% & \textbf{62.9\%} & 39.9\% & \textbf{60.1\%} \\

\midrule
 & GraphRAG\textsubscript{1} & SceneRAG & GraphRAG\textsubscript{1} & SceneRAG & GraphRAG\textsubscript{1} & SceneRAG & GraphRAG\textsubscript{1} & SceneRAG \\
\midrule
\raggedright Comprehensiveness & 39.9\% & \textbf{60.1\%} & 42.1\% & \textbf{57.9\%} & 42.4\% & \textbf{57.6\%} & 40.8\% & \textbf{59.2\%} \\
\raggedright Empowerment       & 37.9\% & \textbf{62.1\%} & 39.5\% & \textbf{60.5\%} & 40.6\% & \textbf{59.4\%} & 38.7\% & \textbf{61.3\%} \\
\raggedright Trustworthiness   & 37.5\% & \textbf{62.5\%} & 36.8\% & \textbf{63.2\%} & 39.7\% & \textbf{60.3\%} & 37.8\% & \textbf{62.2\%} \\
\raggedright Depth             & 37.1\% & \textbf{62.9\%} & 40.4\% & \textbf{59.6\%} & 40.6\% & \textbf{59.4\%} & 38.4\% & \textbf{61.6\%} \\
\raggedright Density           & 34.7\% & \textbf{65.3\%} & 40.8\% & \textbf{59.2\%} & 41.5\% & \textbf{58.5\%} & 37.1\% & \textbf{62.9\%} \\
\raggedright Overall Winner    & 39.0\% & \textbf{61.0\%} & 42.1\% & \textbf{57.9\%} & 42.0\% & \textbf{58.0\%} & 40.1\% & \textbf{59.9\%} \\

\midrule
 & GraphRAG\textsubscript{2} & SceneRAG & GraphRAG\textsubscript{2} & SceneRAG & GraphRAG\textsubscript{2} & SceneRAG & GraphRAG\textsubscript{2} & SceneRAG \\
\midrule
\raggedright Comprehensiveness & 34.8\% & \textbf{65.2\%} & 42.1\% & \textbf{57.9\%} & 41.5\% & \textbf{58.5\%} & 37.5\% & \textbf{62.5\%} \\
\raggedright Empowerment       & 33.1\% & \textbf{66.9\%} & 39.5\% & \textbf{60.5\%} & 40.2\% & \textbf{59.8\%} & 35.6\% & \textbf{64.4\%} \\
\raggedright Trustworthiness   & 30.7\% & \textbf{69.3\%} & 35.1\% & \textbf{64.9\%} & 32.6\% & \textbf{67.4\%} & 31.9\% & \textbf{68.1\%} \\
\raggedright Depth             & 34.3\% & \textbf{65.7\%} & 38.6\% & \textbf{61.4\%} & 40.2\% & \textbf{59.8\%} & 36.2\% & \textbf{63.8\%} \\
\raggedright Density           & 31.2\% & \textbf{68.8\%} & 41.7\% & \textbf{58.3\%} & 40.2\% & \textbf{59.8\%} & 34.9\% & \textbf{65.1\%} \\
\raggedright Overall Winner    & 34.8\% & \textbf{65.2\%} & 41.7\% & \textbf{58.3\%} & 41.1\% & \textbf{58.9\%} & 37.3\% & \textbf{62.7\%} \\

\midrule
 & LightRAG & SceneRAG & LightRAG & SceneRAG & LightRAG & SceneRAG & LightRAG & SceneRAG \\
\midrule
\raggedright Comprehensiveness & 37.9\% & \textbf{62.1\%} & 36.4\% & \textbf{63.6\%} & 37.1\% & \textbf{62.9\%} & 37.5\% & \textbf{62.5\%} \\
\raggedright Empowerment       & 34.7\% & \textbf{65.3\%} & 33.3\% & \textbf{66.7\%} & 33.9\% & \textbf{66.1\%} & 34.3\% & \textbf{65.7\%} \\
\raggedright Trustworthiness   & 35.2\% & \textbf{64.8\%} & 30.3\% & \textbf{69.7\%} & 33.9\% & \textbf{66.1\%} & 34.1\% & \textbf{65.9\%} \\
\raggedright Depth             & 35.2\% & \textbf{64.8\%} & 32.5\% & \textbf{67.5\%} & 36.2\% & \textbf{63.8\%} & 34.9\% & \textbf{65.1\%} \\
\raggedright Density           & 34.0\% & \textbf{66.0\%} & 38.2\% & \textbf{61.8\%} & 33.0\% & \textbf{67.0\%} & 34.6\% & \textbf{65.4\%} \\
\raggedright Overall Winner    & 37.5\% & \textbf{62.5\%} & 36.0\% & \textbf{64.0\%} & 37.1\% & \textbf{62.9\%} & 37.1\% & \textbf{62.9\%} \\

\midrule
& VideoRAG & SceneRAG & VideoRAG & SceneRAG & VideoRAG & SceneRAG & VideoRAG & SceneRAG \\
\midrule
\raggedright Comprehensiveness & 46.1\% & \textbf{53.9\%} & 45.2\% & \textbf{54.8\%} & 46.0\% & \textbf{54.0\%} & 45.9\% & \textbf{54.1\%} \\
\raggedright Empowerment       & 45.3\% & \textbf{54.7\%} & 42.5\% & \textbf{57.5\%} & 43.8\% & \textbf{56.2\%} & 44.5\% & \textbf{55.5\%} \\
\raggedright Trustworthiness   & 44.5\% & \textbf{55.5\%} & 39.0\% & \textbf{61.0\%} & 41.1\% & \textbf{58.9\%} & 42.9\% & \textbf{57.1\%} \\
\raggedright Depth             & 45.7\% & \textbf{54.3\%} & 43.9\% & \textbf{56.1\%} & 46.9\% & \textbf{53.1\%} & 45.6\% & \textbf{54.4\%} \\
\raggedright Density           & 45.2\% & \textbf{54.8\%} & 39.9\% & \textbf{60.1\%} & 45.1\% & \textbf{54.9\%} & 44.2\% & \textbf{55.8\%} \\
\raggedright Overall Winner    & 45.9\% & \textbf{54.1\%} & 45.2\% & \textbf{54.8\%} & 46.0\% & \textbf{54.0\%} & 45.8\% & \textbf{54.2\%} \\
\bottomrule
\end{tabular}

\end{table*}

\begin{figure}
    \centering
    \includegraphics[width=1\textwidth]{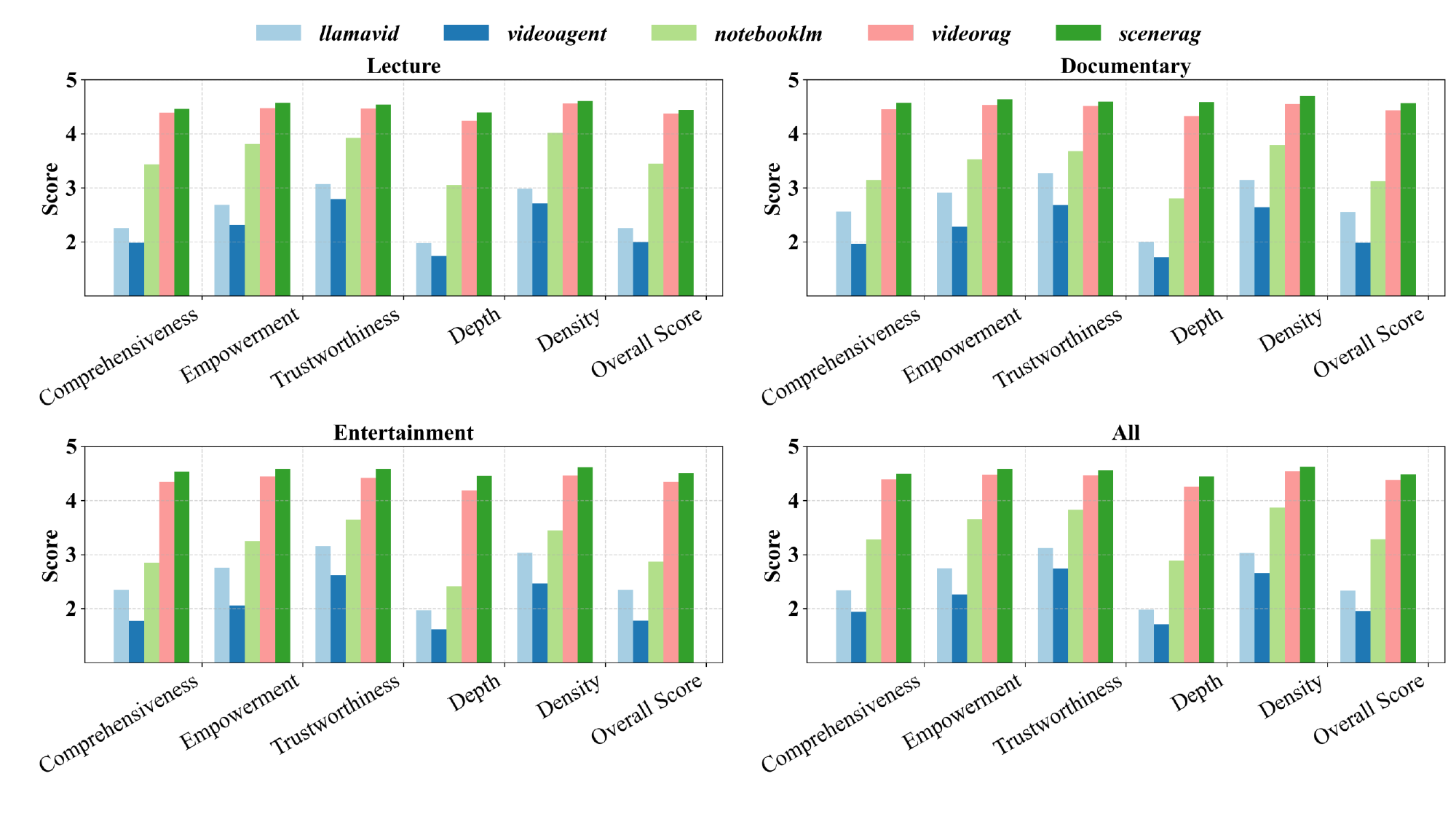} 
    \caption{Quantitative Comparison by gpt-4o-mini. }
    \label{fig:quancompan-4o}
\end{figure}

\begin{table*}[]
\centering
\scriptsize
\caption{Comparison of Winning Rates by GPT-4.1-mini.}
\begin{tabular}{l
  >{\centering\arraybackslash}p{1.0cm}
  >{\centering\arraybackslash}p{1.0cm}
  >{\centering\arraybackslash}p{1.0cm}
  >{\centering\arraybackslash}p{1.0cm}
  >{\centering\arraybackslash}p{1.0cm}
  >{\centering\arraybackslash}p{1.0cm}
  >{\centering\arraybackslash}p{1.0cm}
  >{\centering\arraybackslash}p{1.0cm}
}
\toprule
 & \multicolumn{2}{c}{Lecture} & \multicolumn{2}{c}{Documentary} & \multicolumn{2}{c}{Entertainment} & \multicolumn{2}{c}{All} \\
\cmidrule(lr){2-3} \cmidrule(lr){4-5} \cmidrule(lr){6-7} \cmidrule(lr){8-9}

 & NaiveRAG & SceneRAG & NaiveRAG & SceneRAG & NaiveRAG & SceneRAG & NaiveRAG & SceneRAG \\
\midrule
\raggedright Comprehensiveness & 26.1\% & \textbf{73.9\%} & 23.2\% & \textbf{76.8\%} & 29.9\% & \textbf{70.1\%} & 26.2\% & \textbf{73.8\%} \\
\raggedright Empowerment       & 23.7\% & \textbf{76.3\%} & 19.7\% & \textbf{80.3\%} & 29.0\% & \textbf{71.0\%} & 23.9\% & \textbf{76.1\%} \\
\raggedright Trustworthiness   & 31.5\% & \textbf{68.5\%} & 25.4\% & \textbf{74.6\%} & 37.9\% & \textbf{62.1\%} & 31.6\% & \textbf{68.4\%} \\
\raggedright Depth             & 28.1\% & \textbf{71.9\%} & 22.4\% & \textbf{77.6\%} & 30.8\% & \textbf{69.2\%} & 27.5\% & \textbf{72.5\%} \\
\raggedright Density           & 60.6\% & \textbf{39.4\%} & 57.9\% & \textbf{42.1\%} & 54.5\% & \textbf{45.5\%} & 59.0\% & \textbf{41.0\%} \\
\raggedright Overall Winner    & 27.7\% & \textbf{72.3\%} & 21.1\% & \textbf{78.9\%} & 30.4\% & \textbf{69.6\%} & 26.9\% & \textbf{73.1\%} \\

\midrule
 & GraphRAG\textsubscript{1} & SceneRAG & GraphRAG\textsubscript{1} & SceneRAG & GraphRAG\textsubscript{1} & SceneRAG & GraphRAG\textsubscript{1} & SceneRAG \\
\midrule
\raggedright Comprehensiveness & 20.1\% & \textbf{79.9\%} & 29.4\% & \textbf{70.6\%} & 32.6\% & \textbf{67.4\%} & 24.2\% & \textbf{75.8\%} \\
\raggedright Empowerment       & 11.6\% & \textbf{88.4\%} & 23.2\% & \textbf{76.8\%} & 24.6\% & \textbf{75.4\%} & 16.2\% & \textbf{83.8\%} \\
\raggedright Trustworthiness   & 23.3\% & \textbf{76.7\%} & 23.7\% & \textbf{76.3\%} & 28.6\% & \textbf{71.4\%} & 24.3\% & \textbf{75.7\%} \\
\raggedright Depth             & 17.8\% & \textbf{82.2\%} & 23.2\% & \textbf{76.8\%} & 25.0\% & \textbf{75.0\%} & 20.2\% & \textbf{79.8\%} \\
\raggedright Density           & 35.0\% & \textbf{65.0\%} & 47.8\% & \textbf{52.2\%} & 38.8\% & \textbf{61.2\%} & 38.1\% & \textbf{61.9\%} \\
\raggedright Overall Winner    & 16.1\% & \textbf{83.9\%} & 25.4\% & \textbf{74.6\%} & 26.3\% & \textbf{73.7\%} & 19.8\% & \textbf{80.2\%} \\

\midrule
 & GraphRAG\textsubscript{2} & SceneRAG & GraphRAG\textsubscript{2} & SceneRAG & GraphRAG\textsubscript{2} & SceneRAG & GraphRAG\textsubscript{2} & SceneRAG \\
\midrule
\raggedright Comprehensiveness & 18.5\% & \textbf{81.5\%} & 26.8\% & \textbf{73.2\%} & 34.8\% & \textbf{65.2\%} & 23.1\% & \textbf{76.9\%} \\
\raggedright Empowerment       & 11.7\% & \textbf{88.3\%} & 21.5\% & \textbf{78.5\%} & 29.5\% & \textbf{70.5\%} & 16.9\% & \textbf{83.1\%} \\
\raggedright Trustworthiness   & 20.5\% & \textbf{79.5\%} & 18.0\% & \textbf{82.0\%} & 30.8\% & \textbf{69.2\%} & 21.9\% & \textbf{78.1\%} \\
\raggedright Depth             & 16.4\% & \textbf{83.6\%} & 20.2\% & \textbf{79.8\%} & 30.8\% & \textbf{69.2\%} & 19.8\% & \textbf{80.2\%} \\
\raggedright Density           & 39.0\% & \textbf{61.0\%} & 58.8\% & \textbf{41.2\%} & 60.7\% & \textbf{39.3\%} & 46.8\% & \textbf{53.2\%} \\
\raggedright Overall Winner    & 14.4\% & \textbf{85.6\%} & 20.6\% & \textbf{79.4\%} & 30.4\% & \textbf{69.6\%} & 18.5\% & \textbf{81.5\%} \\

\midrule
 & LightRAG & SceneRAG & LightRAG & SceneRAG & LightRAG & SceneRAG & LightRAG & SceneRAG \\
\midrule
\raggedright Comprehensiveness & 22.6\% & \textbf{77.4\%} & 24.1\% & \textbf{75.9\%} & 25.9\% & \textbf{74.1\%} & 23.5\% & \textbf{76.5\%} \\
\raggedright Empowerment       & 17.2\% & \textbf{82.8\%} & 23.7\% & \textbf{76.3\%} & 21.4\% & \textbf{78.6\%} & 19.2\% & \textbf{80.8\%} \\
\raggedright Trustworthiness   & 27.5\% & \textbf{72.5\%} & 22.8\% & \textbf{77.2\%} & 29.5\% & \textbf{70.5\%} & 27.0\% & \textbf{73.0\%} \\
\raggedright Depth             & 20.1\% & \textbf{79.9\%} & 21.1\% & \textbf{78.9\%} & 22.8\% & \textbf{77.2\%} & 20.8\% & \textbf{79.2\%} \\
\raggedright Density           & 47.1\% & \textbf{52.9\%} & 56.6\% & \textbf{43.4\%} & 48.2\% & \textbf{51.8\%} & 49.1\% & \textbf{50.9\%} \\
\raggedright Overall Winner    & 20.2\% & \textbf{79.8\%} & 20.6\% & \textbf{79.4\%} & 23.2\% & \textbf{76.8\%} & 20.8\% & \textbf{79.2\%} \\

\midrule
& VideoRAG & SceneRAG & VideoRAG & SceneRAG & VideoRAG & SceneRAG & VideoRAG & SceneRAG \\
\midrule
\raggedright Comprehensiveness & 38.7\% & \textbf{61.3\%} & 36.8\% & \textbf{63.2\%} & 42.0\% & \textbf{58.0\%} & 39.0\% & \textbf{61.0\%} \\
\raggedright Empowerment       & 36.2\% & \textbf{63.8\%} & 34.2\% & \textbf{65.8\%} & 39.7\% & \textbf{60.3\%} & 36.5\% & \textbf{63.5\%} \\
\raggedright Trustworthiness   & 37.8\% & \textbf{62.2\%} & 34.2\% & \textbf{65.8\%} & 38.4\% & \textbf{61.6\%} & 37.2\% & \textbf{62.8\%} \\
\raggedright Depth             & 37.1\% & \textbf{62.9\%} & 32.0\% & \textbf{68.0\%} & 38.8\% & \textbf{61.2\%} & 36.5\% & \textbf{63.5\%} \\
\raggedright Density           & 50.9\% & \textbf{49.1\%} & 53.1\% & \textbf{46.9\%} & 58.0\% & \textbf{42.0\%} & 52.7\% & \textbf{47.3\%} \\
\raggedright Overall Winner    & 37.2\% & \textbf{62.8\%} & 32.5\% & \textbf{67.5\%} & 40.6\% & \textbf{59.4\%} & 37.0\% & \textbf{63.0\%} \\
\bottomrule
\end{tabular}

\end{table*}

\begin{figure}
    \centering
    \includegraphics[width=1\textwidth]{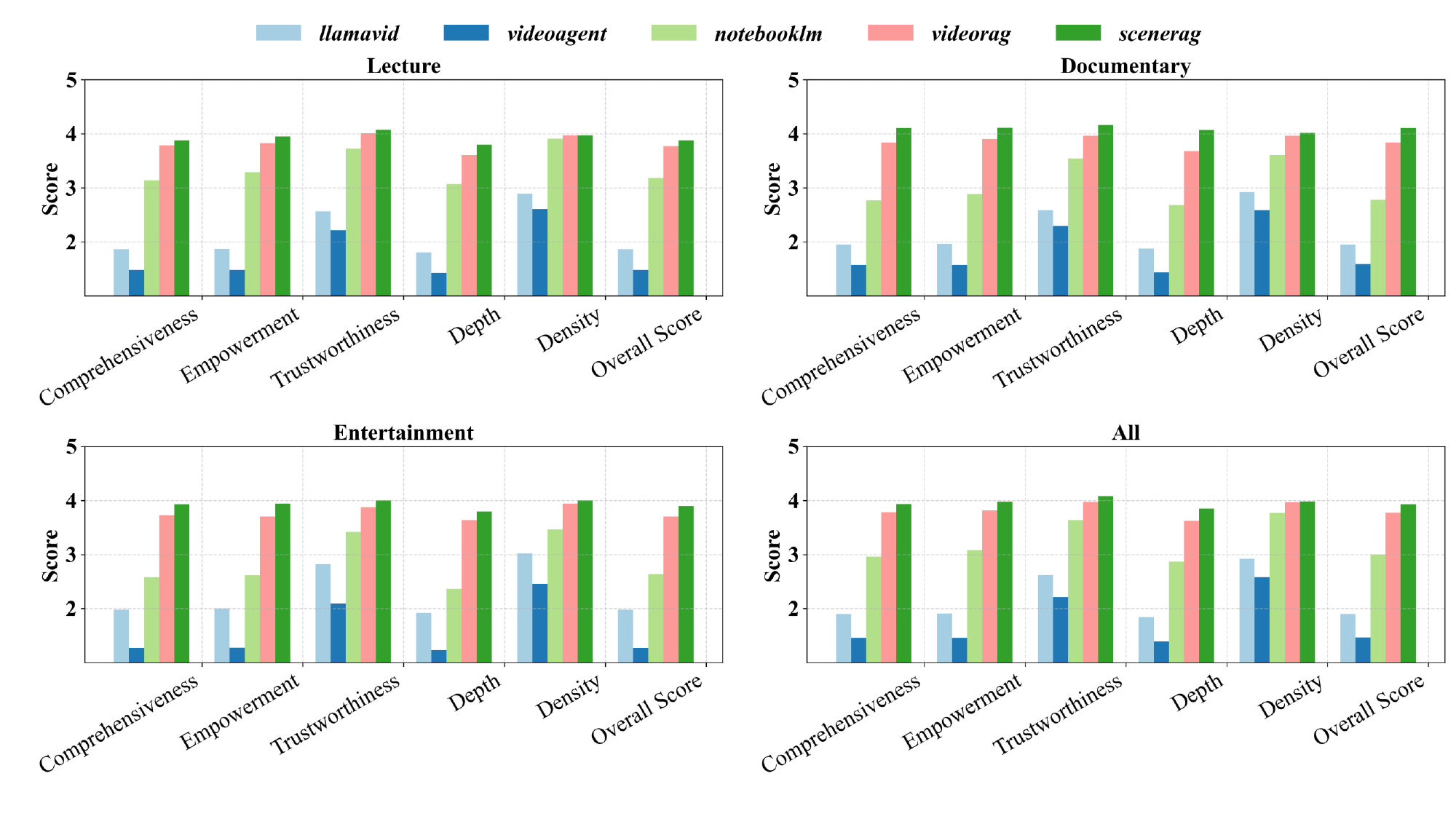} 
    \caption{Quantitative Comparison by gpt-4.1-mini. }
    \label{fig:quancompan-4o}
\end{figure}

\begin{table*}[]
\centering
\scriptsize
\caption{Comparison of Winning Rates by GPT-4.1-nano.}
\begin{tabular}{l
  >{\centering\arraybackslash}p{1.0cm}
  >{\centering\arraybackslash}p{1.0cm}
  >{\centering\arraybackslash}p{1.0cm}
  >{\centering\arraybackslash}p{1.0cm}
  >{\centering\arraybackslash}p{1.0cm}
  >{\centering\arraybackslash}p{1.0cm}
  >{\centering\arraybackslash}p{1.0cm}
  >{\centering\arraybackslash}p{1.0cm}
}
\toprule
 & \multicolumn{2}{c}{Lecture} & \multicolumn{2}{c}{Documentary} & \multicolumn{2}{c}{Entertainment} & \multicolumn{2}{c}{All} \\
\cmidrule(lr){2-3} \cmidrule(lr){4-5} \cmidrule(lr){6-7} \cmidrule(lr){8-9}

 & NaiveRAG & SceneRAG & NaiveRAG & SceneRAG & NaiveRAG & SceneRAG & NaiveRAG & SceneRAG \\
\midrule
\raggedright Comprehensiveness & 39.5\% & \textbf{60.5\%} & 32.5\% & \textbf{67.5\%} & 37.9\% & \textbf{62.1\%} & 37.9\% & \textbf{62.1\%} \\
\raggedright Empowerment       & 39.2\% & \textbf{60.8\%} & 30.3\% & \textbf{69.7\%} & 37.1\% & \textbf{62.9\%} & 37.1\% & \textbf{62.9\%} \\
\raggedright Trustworthiness   & 39.8\% & \textbf{60.2\%} & 28.1\% & \textbf{71.9\%} & 38.8\% & \textbf{61.2\%} & 37.4\% & \textbf{62.6\%} \\
\raggedright Depth             & 38.0\% & \textbf{62.0\%} & 26.3\% & \textbf{73.7\%} & 34.8\% & \textbf{65.2\%} & 35.2\% & \textbf{64.8\%} \\
\raggedright Density           & 50.0\% & \textbf{50.0\%} & 56.1\% & \textbf{43.9\%} & 53.6\% & \textbf{46.4\%} & 51.8\% & \textbf{48.2\%} \\
\raggedright Overall Winner    & 39.0\% & \textbf{61.0\%} & 29.4\% & \textbf{70.6\%} & 36.2\% & \textbf{63.8\%} & 36.6\% & \textbf{63.4\%} \\

\midrule
 & GraphRAG\textsubscript{1} & SceneRAG & GraphRAG\textsubscript{1} & SceneRAG & GraphRAG\textsubscript{1} & SceneRAG & GraphRAG\textsubscript{1} & SceneRAG \\
\midrule
\raggedright Comprehensiveness & 38.8\% & \textbf{61.2\%} & 41.7\% & \textbf{58.3\%} & 43.3\% & \textbf{56.7\%} & 40.2\% & \textbf{59.8\%} \\
\raggedright Empowerment       & 36.2\% & \textbf{63.8\%} & 38.2\% & \textbf{61.8\%} & 39.3\% & \textbf{60.7\%} & 37.1\% & \textbf{62.9\%} \\
\raggedright Trustworthiness   & 34.2\% & \textbf{65.8\%} & 38.2\% & \textbf{61.8\%} & 37.5\% & \textbf{62.5\%} & 35.5\% & \textbf{64.5\%} \\
\raggedright Depth             & 32.7\% & \textbf{67.3\%} & 37.3\% & \textbf{62.7\%} & 35.3\% & \textbf{64.7\%} & 34.1\% & \textbf{65.9\%} \\
\raggedright Density           & 39.2\% & \textbf{60.8\%} & 48.7\% & \textbf{51.3\%} & 45.5\% & \textbf{54.5\%} & 42.2\% & \textbf{57.8\%} \\
\raggedright Overall Winner    & 33.5\% & \textbf{66.5\%} & 37.7\% & \textbf{62.3\%} & 36.6\% & \textbf{63.4\%} & 34.9\% & \textbf{65.1\%} \\

\midrule
 & GraphRAG\textsubscript{2} & SceneRAG & GraphRAG\textsubscript{2} & SceneRAG & GraphRAG\textsubscript{2} & SceneRAG & GraphRAG\textsubscript{2} & SceneRAG \\
\midrule
\raggedright Comprehensiveness & 36.4\% & \textbf{63.6\%} & 38.2\% & \textbf{61.8\%} & 43.8\% & \textbf{56.2\%} & 38.1\% & \textbf{61.9\%} \\
\raggedright Empowerment       & 33.5\% & \textbf{66.5\%} & 36.4\% & \textbf{63.6\%} & 42.9\% & \textbf{57.1\%} & 35.8\% & \textbf{64.2\%} \\
\raggedright Trustworthiness   & 27.5\% & \textbf{72.5\%} & 31.1\% & \textbf{68.9\%} & 30.8\% & \textbf{69.2\%} & 28.8\% & \textbf{71.2\%} \\
\raggedright Depth             & 26.9\% & \textbf{73.1\%} & 31.6\% & \textbf{68.4\%} & 33.5\% & \textbf{66.5\%} & 29.0\% & \textbf{71.0\%} \\
\raggedright Density           & 36.4\% & \textbf{63.6\%} & 53.5\% & \textbf{46.5\%} & 48.7\% & \textbf{51.3\%} & 41.9\% & \textbf{58.1\%} \\
\raggedright Overall Winner    & 28.9\% & \textbf{71.1\%} & 36.0\% & \textbf{64.0\%} & 37.1\% & \textbf{62.9\%} & 31.7\% & \textbf{68.3\%} \\

\midrule
 & LightRAG & SceneRAG & LightRAG & SceneRAG & LightRAG & SceneRAG & LightRAG & SceneRAG \\
\midrule
\raggedright Comprehensiveness & 36.7\% & \textbf{63.3\%} & 35.1\% & \textbf{64.9\%} & 37.5\% & \textbf{62.5\%} & 36.5\% & \textbf{63.5\%} \\
\raggedright Empowerment       & 33.6\% & \textbf{66.4\%} & 33.3\% & \textbf{66.7\%} & 35.7\% & \textbf{64.3\%} & 34.0\% & \textbf{66.0\%} \\
\raggedright Trustworthiness   & 31.8\% & \textbf{68.2\%} & 32.9\% & \textbf{67.1\%} & 32.1\% & \textbf{67.9\%} & 32.1\% & \textbf{67.9\%} \\
\raggedright Depth             & 30.9\% & \textbf{69.1\%} & 29.8\% & \textbf{70.2\%} & 29.9\% & \textbf{70.1\%} & 30.5\% & \textbf{69.5\%} \\
\raggedright Density           & 46.5\% & \textbf{53.5\%} & 56.1\% & \textbf{43.9\%} & 47.3\% & \textbf{52.7\%} & 48.5\% & \textbf{51.5\%} \\
\raggedright Overall Winner    & 31.6\% & \textbf{68.4\%} & 32.0\% & \textbf{68.0\%} & 33.0\% & \textbf{67.0\%} & 32.0\% & \textbf{68.0\%} \\

\midrule
& VideoRAG & SceneRAG & VideoRAG & SceneRAG & VideoRAG & SceneRAG & VideoRAG & SceneRAG \\
\midrule
\raggedright Comprehensiveness & 46.7\% & \textbf{53.3\%} & 38.6\% & \textbf{61.4\%} & 42.0\% & \textbf{58.0\%} & 44.3\% & \textbf{55.7\%} \\
\raggedright Empowerment       & 46.3\% & \textbf{53.7\%} & 40.4\% & \textbf{59.6\%} & 43.8\% & \textbf{56.2\%} & 44.7\% & \textbf{55.3\%} \\
\raggedright Trustworthiness   & 44.9\% & \textbf{55.1\%} & 39.0\% & \textbf{61.0\%} & 40.6\% & \textbf{59.4\%} & 43.0\% & \textbf{57.0\%} \\
\raggedright Depth             & 44.7\% & \textbf{55.3\%} & 39.0\% & \textbf{61.0\%} & 41.1\% & \textbf{58.9\%} & 42.9\% & \textbf{57.1\%} \\
\raggedright Density           & 46.9\% & \textbf{53.1\%} & 48.7\% & \textbf{51.3\%} & 49.6\% & \textbf{50.4\%} & 47.8\% & \textbf{52.2\%} \\
\raggedright Overall Winner    & 45.2\% & \textbf{54.8\%} & 39.0\% & \textbf{61.0\%} & 41.5\% & \textbf{58.5\%} & 43.4\% & \textbf{56.6\%} \\
\bottomrule
\end{tabular}
\end{table*}

\begin{figure}
    \centering
    \includegraphics[width=1\textwidth]{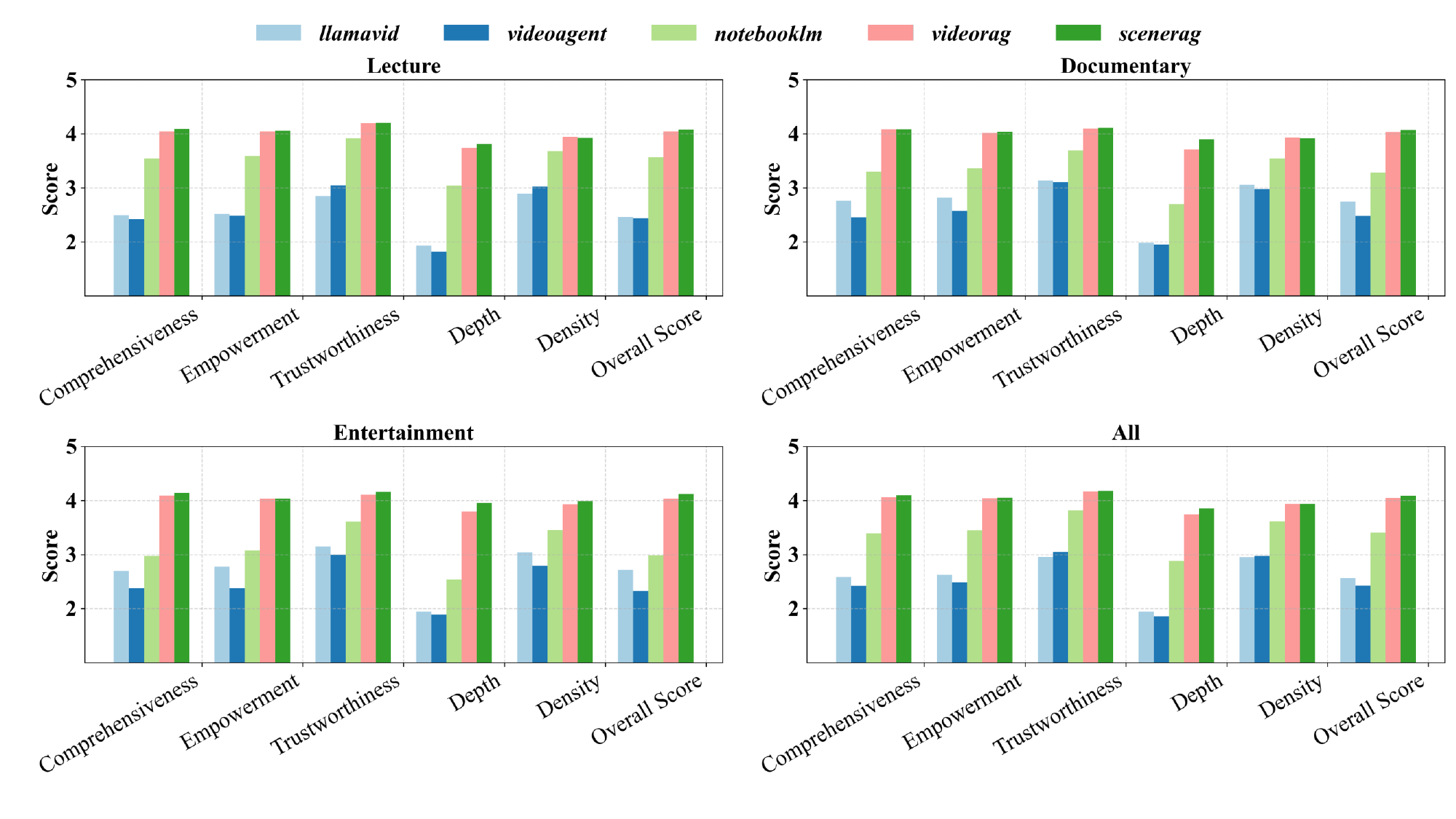} 
    \caption{Quantitative Comparison by gpt-4.1-nano. }
    \label{fig:quancompan-4.1nano}
\end{figure}

\clearpage

\section{Dataset Descriptions}
The LongerVideos dataset~\cite{ren2025videorag} is designed to evaluate models on long-form video comprehension and knowledge extraction. It comprises diverse video lists sourced primarily from YouTube, including online courses and thematic compilations, with durations ranging from just a few minutes to several hours. Each list is paired with a set of open-ended questions that require integrating information across multiple videos. Videos were collected using yt-dlp and questions were curated with assistance from advanced multi-video QA tools. In total, the dataset contains 22 curated video lists, facilitating comprehensive assessment of a model's ability to synthesize, reason, and provide accurate responses over extended video content. Detailed dataset statistics can be found in Table~\ref{tab:longervideos-stats}.

\begin{table*}[!htbp]
\centering
\caption{Detailed statistics of the \textit{LongerVideos} dataset.}
\label{tab:longervideos-stats}
\begin{tabular}{llccc}
\toprule
\textbf{Video Type} & \textbf{video list name} & \textbf{\#video} & \textbf{\#query} & \textbf{overall duration} \\
\midrule
\multirow{12}{*}{Lecture}
    & climate-week-at-columbia-engineering     & 4  & 26 & 5.91 hours \\
    & rag-lecture                             & 19 & 38 & 5.34 hours \\
    & ai-agent-lecture                        & 39 & 45 & 9.35 hours \\
    & daubechies-wavelet-lecture              & 4  & 25 & 8.97 hours \\
    & daubechies-art-and-mathematics-lecture  & 4  & 21 & 4.87 hours \\
    & tech-ceo-lecture                        & 4  & 31 & 4.83 hours \\
    & dspy-lecture                            & 9  & 38 & 4.22 hours \\
    & trading-for-beginners                   & 2  & 23 & 4.11 hours \\
    & ahp-superdecision                       & 11 & 24 & 2.40 hours \\
    & decision-making-science                 & 4  & 26 & 2.20 hours \\
    & 12-days-of-openai                       & 12 & 35 & 3.43 hours \\
    & autogen                                 & 23 & 44 & 8.70 hours \\
\midrule
\multirow{5}{*}{Documentary}
    & fights-in-animal-kingdom                & 1 & 11 & 3.00 hours \\
    & nature-scenes                           & 1 & 17 & 3.98 hours \\
    & education-united-nations                & 6 & 39 & 8.41 hours \\
    & elon-musk                               & 1 & 13 & 8.63 hours \\
    & jeff-bezos                              & 3 & 34 & 4.47 hours \\
\midrule
\multirow{5}{*}{Entertainment}
    & black-myth-wukong                       & 10 & 23 & 21.36 hours \\
    & primetime-emmy-awards                   & 3  & 17 & 7.31 hours \\
    & journey-through-china                   & 1  & 27 & 3.37 hours \\
    & fia-awards                              & 1  & 27 & 3.02 hours \\
    & game-awards                             & 2  & 18 & 6.73 hours \\
\bottomrule
\end{tabular}
\end{table*}

\section{Details of Case Study}

To evaluate SceneRAG’s effectiveness in segmented retrieval and synthesis, we conducted a case study using the long-form technical lecture \textit{“Is This the End of RAG? Anthropic’s NEW Prompt Caching”}. The core query—“How does prompt caching compare to traditional RAG in terms of cost and efficiency?”—requires aggregating evidence distributed across the video timeline. SceneRAG’s discourse-aware segmentation algorithm identified two non-contiguous yet complementary scenes ([23.96–66.08,s] and [213.12–292.48,s]), which together serve as a comprehensive and focused basis for answering the query, as summarized in Table~\ref{tab:scene_segments}. The first segment introduces the motivation and cost/latency benefits of prompt caching, while the second presents quantitative comparisons (e.g., “up to 80–90\% reduction in both latency and cost”) and a trade-off analysis between Anthropic and Gemini implementations, detailed in Table~\ref{tab:scene_caption_transcript}.

For transparency, SceneRAG highlights key evidence at specific timestamps—\textcolor{green!60!black}{24,s}, \textcolor{cyan!80!black}{40,s}, \textcolor{purple}{214,s}, and \textcolor{red!60!black}{275,s}—which directly correspond to the main arguments in the retrieved scenes. This approach ensures that the synthesized answer is both contextually rich and easily traceable to the original material. The case study thus demonstrates SceneRAG’s advantage in precise, context-preserving retrieval for complex technical queries.

\begin{table}[htbp]
\caption{A case study evaluating SceneRAG's segmented retrieval and synthesis capabilities on long-form video. The system is tested on the lecture ``Is This the End of RAG? Anthropic’s NEW Prompt Caching,'' using a query focused on cost and efficiency comparisons between prompt caching and traditional RAG. Color legend: \textcolor{green!60!black}{green} = Cost-related; \textcolor{cyan!80!black}{Blue} = Efficiency/latency-related.}
\renewcommand{\arraystretch}{1.2}
\setlength{\arrayrulewidth}{1pt}  

\begin{tabular}{|p{0.97\textwidth}|}
\hline
\rowcolor{white}
\vspace{0.1cm}
\begin{tcolorbox}[colback=white,colframe=black,boxrule=0.6pt,left=0pt,right=0pt,top=1pt,bottom=1pt]
\textbf{Query:} \textbf{How does prompt caching compare to traditional RAG in terms of cost and efficiency?}
\end{tcolorbox} \\

\begin{tcolorbox}[colback=white,left=0pt,right=0pt,top=1pt,bottom=1pt,boxrule=0.6pt]
\textbf{Input Video:} \textcolor{red}{\textbf{Is This the End of RAG? Anthropic’s NEW Prompt Caching} (Lecture)}
\end{tcolorbox} \\

\textbf{SceneRAG Response:} \\

\#\#\# Overview of Prompt Caching vs. Traditional RAG\\

Prompt caching and traditional Retrieval-Augmented Generation (RAG) are both techniques designed to improve the efficiency and effectiveness of language models. However, they differ significantly in terms of cost efficiency, processing time, and usage scenarios.

Cost Efficiency\\

\textcolor{green!60!black}{One of the primary advantages of prompt caching is its remarkable cost-effectiveness. For instance, prompt caching can reduce costs by up to 90\% compared to traditional methods when dealing with extensive prompts or multiple interactions. This reduction is attributed to the fact that cached tokens cost significantly less---approximately 10\% of the base input token price---thus leading to substantial savings during multiple API calls (Source: Video on Prompt Caching).
In comparison, traditional RAG systems can incur higher costs, especially when a large number of tokens or documents are fetched repeatedly from a database. Although RAG systems allow for the retrieval of only the most relevant information based on specific queries, the contextual embeddings and processing involved can lead to increased operational expenses over time, mainly due to the need for constant storage and retrieval of data (Source: Video on RAG systems).}

Efficiency and Processing Time\\

\textcolor{cyan!80!black}{When it comes to latency and processing times, prompt caching also shows impressive results. For example, when handling prompts of 100000 tokens, the response time can drop significantly from about 12 seconds without caching to approximately 2.4 seconds with caching---representing an 80\% reduction in processing time. Similarly, even in many-shot prompting scenarios, prompt caching demonstrates a latency reduction of about 31\% (Source: Videos detailing prompt caching benefits).
Traditional RAG methods tend to have longer response times. RAG typically requires an initial retrieval step, which can introduce delays, particularly when large datasets are involved. The complexity involved in retrieving, processing, and generating coherent responses from various chunks of data can result in prolonged wait times for end-users (Source: Video on RAG systems).}
(The remaining content are omitted for brevity.)

\begin{tcolorbox}[colback=white,boxrule=0.6pt,left=0pt,right=0pt,top=1pt,bottom=1pt]
SceneRAG precisely retrieved two non-contiguous, highly relevant video segments from the lecture \href{https://www.youtube.com/watch?v=Fv_j52DDJUE}{\textcolor{blue}{\textit{Is This the End of RAG? Anthropic’s NEW Prompt Caching}}}, covering both the motivation and technical comparisons between prompt caching and traditional RAG. Here, we highlight the main evidence aligned with the answer at key timestamps: \textcolor{green!60!black}{\textbf{24\,s}}, \textcolor{cyan!80!black}{\textbf{40\,s}}, \textcolor{purple}{\textbf{214\,s}}, and \textcolor{red!60!black}{\textbf{275\,s}}.\\
From left to right, these correspond to: the introduction of prompt caching’s cost/latency benefits (\textcolor{green!60!black}{24\,s}, \textcolor{cyan!80!black}{40\,s}), and concrete quantitative comparisons, including 80--90\% cost and latency reduction and cross-system trade-offs (\textcolor{purple}{214\,s}, \textcolor{red!60!black}{275\,s}). These moments collectively supply a focused, complete answer to the cost and efficiency comparison query.
\end{tcolorbox} \\

\multicolumn{1}{|c|}{
\begin{minipage}{0.97\textwidth}
\centering
\includegraphics[width=0.4\textwidth]{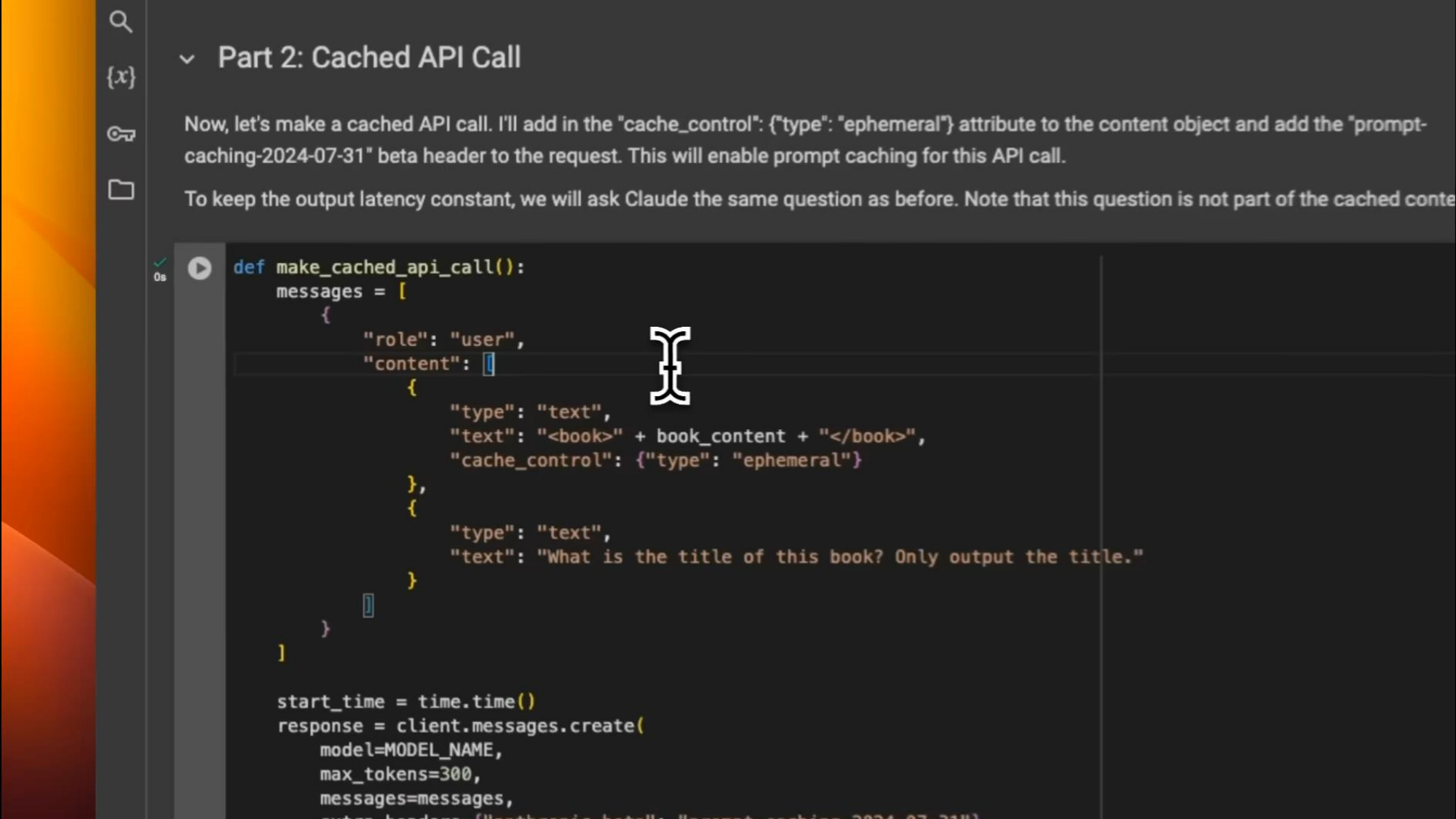}\hspace{1em}
\includegraphics[width=0.4\textwidth]{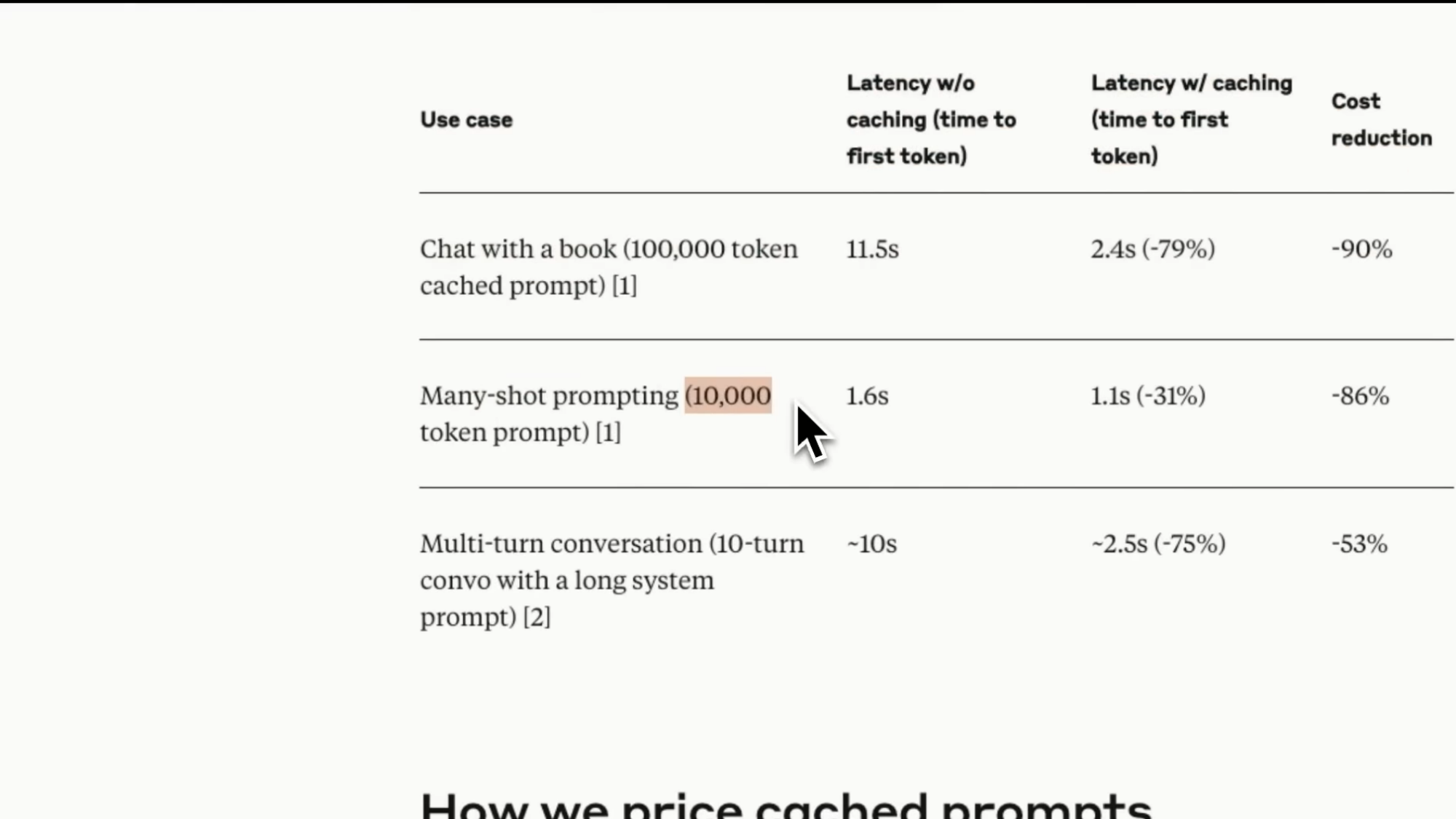}
\vspace{0.6em}  
\end{minipage}
} \\

\hline
\end{tabular}
\end{table}

\begin{table}
\centering
\caption{Segmented Scenes and Transcripts in the Video}
\label{tab:scene_segments}
\begin{tabular}{p{13.5cm}}
\toprule
\textbf{Scene and Transcript Content} \\
\midrule
\textbf{[0.00s,23.96s]} \newline
So Anthropic just introduced prompt caching with Cloud. That can reduce cost up to by 90\% and latency up to by 85, which is huge. And did they just kill Rag with this new feature? Now Google was the first one who introduced context caching with their Gemini models. There are some similarities, but some major differences as well between these two approaches. We will discuss them later in the video. I'll show you how to get started. And what type of performance difference you can expect. Before looking at the quote example, let's see what was released today. Prompt caching enables developers to cache frequently used contacts between API calls. \\

\textbf{[23.96s,66.08s]} \newline
Anthropic models have a huge context window of 200,000 tokens. However, if you're chatting with long documents, you'll have to send them with each prompt. So that becomes very expensive. And hence, the prompt caching is going to be extremely helpful. So now customers can provide Cloud with more background information and example outputs or few short prompting. Reducing cost by up to 90\% and latency up to 85\%. Now, these numbers are great, but they are not going to be consistent based on the example use cases. And we are going to look at some of them. This feature is available both for Cloud 3.5 Sonnet and Cloud 3 Haiku. Support for Cloud 3 opus is coming soon. \\

\textbf{[66.08s,134.22s]} \newline
As I said in the beginning, context caching has been available for the Gemini models. And there are some interesting differences between the two, which I'm going to highlight throughout the video. So what are going to be some use cases for prompt caching? Well, the first one is conversational agents. So if you're having a long-form conversation and there is a substantial chat history, you can put that chat history in the cache and just ask questions from that. Another example, use case is coding assistants. Usually code bases are pretty huge. So you can put them in the prompt cache and then use your subsequent messages for question answer. Also, launch document processing or detailed instruction sets. This specifically will apply if you have highly detailed system prompt with a lot of few short examples. So this is going to be very helpful that you can just send those ones and then you can have subsequent conversations while this is cached. \\

\textbf{[134.22s,213.12s]} \newline
A genetic search and tool usage is another example, especially if you have to define your tools, what inputs are to different tools so you can put them in your prompt cache and then send that once and that will save you a lot of money. And that example is going to be talked to books, papers, documentation, podcast, transcripts and other long form content. So this is a very enticing application for Rack and with these long context models, especially with prompt caching or context caching, now it becomes viable to just put these documents in the context rather than chunking them, computing embedding, and then doing retrieval on the documents. Now here's a table that shows what type of reduction in cost and latency you can expect for different applications. \\

\textbf{[213.12s,300.00s]} \newline
If you're chatting with your documents and you're sending 100,000 tokens without caching, that would take about 12 seconds to generate a response, but with caching, that's about only 2.4 or 2.5 seconds, which is 80\% reduction in processing time on latency and 90\% reduction in the cost. If you're doing a few shot prompting with 10,000 tokens, you expect about 31\% reduction in latency and about 86\% reduction in cost. Whereas if you're doing multi-turn conversation, a 10-turn conversation, you're expecting about 75\% reduction in latency, but only about 53\% reduction in cost. Now, the way the cash to tokens are charged versus the input output tokens are different and that's why you see these reductions in the cost as well. Now, we saw the cost reduction because the cash tokens are costing only 10\% of the base input token price, which is a huge reduction of 90\%. However, you also need to keep in mind that writing to the cash costs about 25\% more than the base input token price for any given model. So there is an overhead when you have a writing to the cash for the first time, but then there is a substantial reduction in cost. Now, the Gemini models do it in a different way. There is no cost associated with the actual cash token, but there is a storage cost of \$1 per million tokens per hour. Okay, so here's what the reduction is going to look like. \\
\bottomrule
\end{tabular}
\end{table}

\begin{table}
\centering
\caption{Scene Transcripts and Captions of the Video}
\label{tab:scene_caption_transcript}
\begin{tabular}{p{6.05cm} p{6.95cm}}
\toprule
\textbf{Scene and Transcript Content} & \textbf{Scene Description (Caption)} \\
\midrule
\textbf{[23.96s,66.08s]} \newline
Anthropic models have a huge context window of 200,000 tokens. However, if you're chatting with long documents, you'll have to send them with each prompt. So that becomes very expensive. And hence, the prompt caching is going to be extremely helpful. So now customers can provide Cloud with more background information and example outputs or few short prompting. Reducing cost by up to 90\% and latency up to 85\%. Now, these numbers are great, but they are not going to be consistent based on the example use cases. And we are going to look at some of them. This feature is available both for Cloud 3.5 Sonnet and Cloud 3 Haiku. Support for Cloud 3 opus is coming soon. 
& 
The video presents a tutorial on using prompt caching with Anthropic's AI models, focusing particularly on reducing costs and improving response latency. It begins by showcasing the integration of prompt caching into an API call script within a development environment, highlighting the use of \texttt{cache\_control} headers to enable this feature. The video then transitions to a web page detailing how prompt caching works, its benefits, and its availability in public beta for specific versions of the models (Claude 3 Sonnet and Claude 3 Haiku). As the narration explains these points, the webpage is displayed with clear visuals such as icons and text that support the explanation. The video emphasizes the practical applications of prompt caching, including its effectiveness in situations requiring large amounts of context, enhancing background knowledge, and optimizing performance through reduced costs and faster responses. Throughout, the focus remains on educating viewers about the implementation and advantages of prompt caching without showing any human subjects or interactions beyond the digital interface being demonstrated. \\

\textbf{[213.12s,300.00s]} \newline
If you're chatting with your documents and you're sending 100,000 tokens without caching, that would take about 12 seconds to generate a response, but with caching, that's about only 2.4 or 2.5 seconds, which is 80\% reduction in processing time on latency and 90\% reduction in the cost. If you're doing a few shot prompting with 10,000 tokens, you expect about 31\% reduction in latency and about 86\% reduction in cost. Whereas if you're doing multi-turn conversation, a 10-turn conversation, you're expecting about 75\% reduction in latency, but only about 53\% reduction in cost. Now, the way the cash to tokens are charged versus the input output tokens are different and that's why you see these reductions in the cost as well. Now, we saw the cost reduction because the cash tokens are costing only 10\% of the base input token price, which is a huge reduction of 90\%. However, you also need to keep in mind that writing to the cash costs about 25\% more than the base input token price for any given model. So there is an overhead when you have a writing to the cash for the first time, but then there is a substantial reduction in cost. Now, the Gemini models do it in a different way. There is no cost associated with the actual cash token, but there is a storage cost of \$1 per million tokens per hour. Okay, so here's what the reduction is going to look like. 
& 
The video presents a detailed comparison of prompt caching in natural language processing, focusing on the reduction in latency and cost when using cached content. It begins by displaying a table with three use cases: chatting with documents, many-shot prompting, and multi-turn conversations, each showing reduced latency times ranging from 12 seconds to over 30 seconds without caching, down to approximately 2.5 seconds with caching, resulting in up to an 86\% cost reduction. The visual transitions smoothly through various sections of text, highlighting key points about pricing based on token counts, input/output tokens, and storage costs. The video explains that writing to cache incurs additional charges, but using cached content significantly lowers these costs, often costing only 10\% of the base input token price. This information is reinforced through a dark-themed interface showcasing rate limits, context caching, and output prices for different models like Claude 3.5 Sonnet, Claude 3 Opus, and Gemini models. Each model's details are presented with specific pricing structures, emphasizing how the cost varies depending on factors such as the number of tokens, input/output prompts, and storage requirements. Throughout the video, the cursor moves across the screen, pointing out significant figures and changes in the text, ensuring viewers understand the financial implications of prompt caching. The consistent transition between frames keeps the focus on the educational content, providing a clear overview of how prompt caching can optimize performance while minimizing expenses. \\
\bottomrule
\end{tabular}
\end{table}

\end{document}